\documentclass[10pt,twocolumn,letterpaper]{article}

 \usepackage[pagenumbers]{cvpr} 

\usepackage{soul}

\usepackage{array}
\usepackage{relsize}
\usepackage{textcomp}

\usepackage[utf8]{inputenc}
\DeclareUnicodeCharacter{2033}{''}

\definecolor{cvprblue}{rgb}{0.21,0.49,0.74}
\usepackage[pagebackref,breaklinks,colorlinks,allcolors=cvprblue]{hyperref}

\usepackage{xcolor}
\usepackage{soul}
\usepackage{algorithm}
\usepackage{algorithmic}
\usepackage{graphicx}
\usepackage{rotating}

\def\paperID{N/A} 
\def\confName{CVPR}
\def\confYear{2026}

\title{SpiderCam: Low-Power Snapshot Depth from Differential Defocus}

\author{Marcos A. Ferreira$^{*,1}$ \quad Tianao Li$^{*,1}$ \quad John Mamish$^{*,2}$ \\ Josiah Hester$^{2}$ \quad Yaman Sangar$^{2}$ \quad Qi Guo$^{3}$ \quad Emma Alexander$^{1}$ \\
$^{1}$Northwestern University \quad
$^{2}$Georgia Institute of Technology \quad
$^{3}$Purdue University \\ 
$^{*}$Equal contributions, {\tt\small 
\href{https://nubivlab.github.io/SpiderCam}{https://nubivlab.github.io/SpiderCam}
}
}

\begin{document}

\maketitle
\begin{abstract}
We introduce SpiderCam, an FPGA-based snapshot depth-from-defocus camera which produces 480$\times$400 sparse depth maps in real-time at 32.5 FPS over a working range of 52 cm while consuming 624 mW of power in total. SpiderCam comprises a custom camera that simultaneously captures two differently focused images of the same scene, processed with a SystemVerilog implementation of depth from differential defocus (DfDD) on a low-power FPGA. To achieve state-of-the-art power consumption, we present algorithmic improvements to DfDD that overcome challenges caused by low-power sensors, and design a memory-local implementation for streaming depth computation on a device that is too small to store even a single image pair. We report the first sub-Watt total power measurement for passive FPGA-based 3D cameras in the literature.

\end{abstract}

\section{Introduction}
\label{sec:intro}

Jumping spiders accurately estimate distances using a brain the size of a poppy seed. Their behavior suggests that they use depth from defocus~\cite{nagata2012depth}, aided by an eye that captures pairs of differently-defocused images simultaneously across layered retinas~\cite{land1969structure}. Inspired by their example, depth from differential defocus (DfDD) is a computationally efficient class of depth from defocus (DfD) algorithms that leverages small changes in defocus~\cite{alexander2019theory}. While DfDD has been shown to be more efficient than other DfD approaches in terms of floating point operations (FLOPs)~\cite{luo2025depth}, it has not previously been demonstrated in a real-world ultra-low-power system.

Low power depth sensing has instead been dominated by stereo, with the most efficient FPGA-based prototypes demonstrating 2-3 W of overall power consumption~\cite{mattoccia2015passive,puglia2017real,ttofis2015low,wang2022block}. Significantly, many existing works don't measure real-world power consumption or report quantitative real-world depth results (see \cref{tab:powertable}). Instead, they benchmark on high-quality datasets like KITTI~\cite{Geiger2012CVPR,Geiger2013IJRR,Fritsch2013ITSC,Menze2015CVPR} and Middlebury~\cite{scharstein2002taxonomy,scharstein2003high,scharstein2007learning,scharstein2014high,hirschmuller2007evaluation}, and estimate the ``core power" of their algorithm alone. This ignores the power already expended through high-quality image sensors, I/O, and image alignment. It also likely overestimates depth accuracy for highly constrained systems, which may suffer from effects like sensor noise, miscalibration, and small baselines.

We show, for the first time, a complete FPGA-based 3D camera system operating under 1 Watt, and demonstrate a working range of over half a meter on real scenes measured on-device. Our camera captures simultaneous image pairs through a beam splitter with sensors at different offsets, like the eye of the jumping spider. In real time, our device outputs sparse depth maps, filtered by measurement confidence. Importantly, we demonstrate both the real-world total system power (including both sensors) and the real-world working range, on data with nonidealities from a compact, snapshot, low-power camera.

The algorithmic challenges presented by our tight power budget are threefold. First, the total operations are limited, even compared to previous DfDD methods, and we can no longer ignore the relative cost of different floating point operations (e.g., division can be 10$\times$ more expensive than multiplication). Second, low-power FPGAs impose tight resource constraints, requiring a memory-local streaming approach with a limited spatial footprint. Third, low-power sensors are small and noisy, requiring us to develop an algorithm with improved robustness to both optical aberrations and numerical errors. By combining a spider-like camera design for compact simultaneous image capture, with a novel algorithm tailored to our low-power electronics, SpiderCam demonstrates the promise of defocus for real-world low-power depth sensing.

\section{Related Work}
\label{sec:related_work}

\begin{table*}[bhtp!]
    \centering
    \resizebox{2\columnwidth}{!}{
    \scriptsize
    \centering
    \begin{tabular}{
        p{2.2cm}  
        >{\centering\arraybackslash}p{0.5cm}
        >{\centering\arraybackslash}p{1.1cm}
        >{\centering\arraybackslash}p{1.2cm}
        >{\centering\arraybackslash}p{0.5cm}
        >{\centering\arraybackslash}p{0.5cm}
        >{\centering\arraybackslash}p{3cm}
        >{\centering\arraybackslash}p{2.5cm}
    }
        \toprule
        \textbf{Name} & \textbf{Type}       & \textbf{FPGA}              & \textbf{Resolution} & 
        \multicolumn{2}{c}{\textbf{Core Power (W)}}
         & \multicolumn{2}{c}{\textbf{Real-World Quantitative Results}}\\ 
                      &                          &                            &                     & Min & Max & On-Device Working Range & Full System Power \\
        \midrule
        
        Jin, 2014 \cite{jin2014fast}    & Stereo            & Kintex-6                & 640$\times$480         &
        1.93& 2.67 & none & none \\

        Raj, 2014 \cite{joseph2014video}  & DfD        & Virtex-4\textsuperscript{\textdagger}            & 400$\times$400        &
        0.46& 0.58   & 0.77--0.80 m (\textbf{0.5$\%$ err}) & 2 W + camera     \\

        Mattoccia, 2015 \cite{mattoccia2015passive} & Stereo          & Spartan-7               & 640$\times$480          &
        0.44& 0.68 & qualitative & 2.5 W              \\

        Ttofis, 2015 \cite{ttofis2015low}    & Stereo            &  Kintex-7               &  1280$\times$720        &
        0.92& 1.53 & qualitative & 2.8 W \\
        
        Puglia, 2017 \cite{puglia2017real}   & Stereo            &  Zynq Artix             &  1024$\times$768       &  
        \underline{0.43} & \underline{0.68} & none & \underline{2 W} \\

        Zhang, 2018 \cite{zhang2018nipm}    & Stereo            & Kintex-7                & 1920$\times$1080        &
        0.89& 1.23 & none & none \\

        Lu, 2021 \cite{lu2021resource}    & Stereo            & Kintex-7                & 1024$\times$480        &
        1.39& 2.30 & none & none \\

        Wang, 2022 \cite{wang2022block}    & Stereo            & Zynq US+                & 1920$\times$1080       &
        2.03& 2.20 & none & 2.8 W + camera        \\
        
        Ours, normalized                     & DfD       & Kintex-7            & 512$\times$480          &
        \textbf{0.42}& \textbf{0.55} & - & -
        \\
         Ours, actual                     & DfD       &  ECP5            & 480$\times$400          &
        \textbf{0.24} & \textbf{0.31}  & \underline{0.45--0.97 m} ($10\%$ err) & \textbf{0.6 W} 
        \\
        \midrule
        Luo, 2025 \cite{luo2025focal} & DfD & N/A & 480$\times$360 & N/A & N/A & \textbf{0.40--1.20 m} ($10\%$ err) & 4.9 W 
        \\
        
        \bottomrule
    \end{tabular}

   }
    \caption{\textbf{Power consumption for passive FPGA-based depth imaging systems.} ``Core Power" numbers reflect only power consumed by logic for depth computation on aligned images; I/O and sensor power are ignored. Power estimates are derived using vendor tools and data from the corresponding works, normalized to a processing rate of 30 Mpix/sec across methods, see ~\cref{sec:power_estimates_supp} for details. In addition to SOTA power efficiency in core power estimates, we are one of the very few systems to report full system power consumption and real-world depth performance (from noisy sensor captures rather than on existing datasets collected with large-baseline, high-power, pre-aligned image pairs). Note that we report two estimates for our method: a ``normalized" estimate that handles more pixels per second on a less efficient FPGA for fairer comparisons with the literature, and a lower estimate reflecting the true parameters of our system. Additionally, we include Focal Split (Luo, 2025) \cite{luo2025focal}, which reports a similar method on a Raspberry Pi 5 with real-world depth and power performance.
    \textsuperscript{\textdagger} This work uses an Xilinx Virtex-2 FPGA, but due to vendor tool unavailability, we estimate power for the more efficient Virtex-4.
    }
    \label{tab:powertable}
\end{table*}

Depth imaging has been a major area of imaging research for decades, with solutions spanning the accuracy, speed, and SWaP-C (size, weight, power, and cost) tradeoff space. Broadly speaking, depth imaging can be separated into two categories: active and passive. Active techniques use a controlled light source to improve signal quality~\cite{giancola2018survey,horaud2016overview}, 
like LiDAR~\cite{behroozpour2017lidar,geng2011structured} and structured light~\cite{geng2011structured, mirdehghan2018optimal, o20143d}. Reducing the SWaP-C of active techniques is challenging due to their need for illumination sources~\cite{giancola2018survey,raj2020survey,lee2020accuracy,realsensemanual}. 
Passive depth imaging systems, on the other hand, calculate depth images without using special hardware to illuminate the scene, and are the focus of this paper.

\subsection{Monocular Depth Imaging}
Estimating depth from a single monocular image is an ill-posed problem and cannot be solved without extensive reliance on priors~\cite{bhoi2019monocular,ming2021deep}. In spite of this, dense, high-accuracy results that run in real-time have been achieved by using deep learning to leverage large datasets. Although successful, these methods are ill-suited to space-constrained mobile computing: they rely on deep neural networks whose size~\cite{ming2021deep} leads to computation and memory access energy costs which prevent the miniaturization of systems that use them, even with state-of-the-art edge TPUs~\cite{shuvo2022efficient,ngo2025edge,gholami2024ai}. If motion is introduced and a sequence of frames is used for the monocular depth imaging problem, it is no longer ill-posed and a broader class of both classical and deep learning algorithms is admitted~\cite{ozyecsil2017survey,masoumian2022monocular}. Although these methods yield high-quality results, their adoption in mobile computing systems is limited due to their compute and memory requirements~\cite{schonberger2016structure,saputra2018visual}.

\begin{figure*}[ht]
    \centering
    \includegraphics[width=1\linewidth]{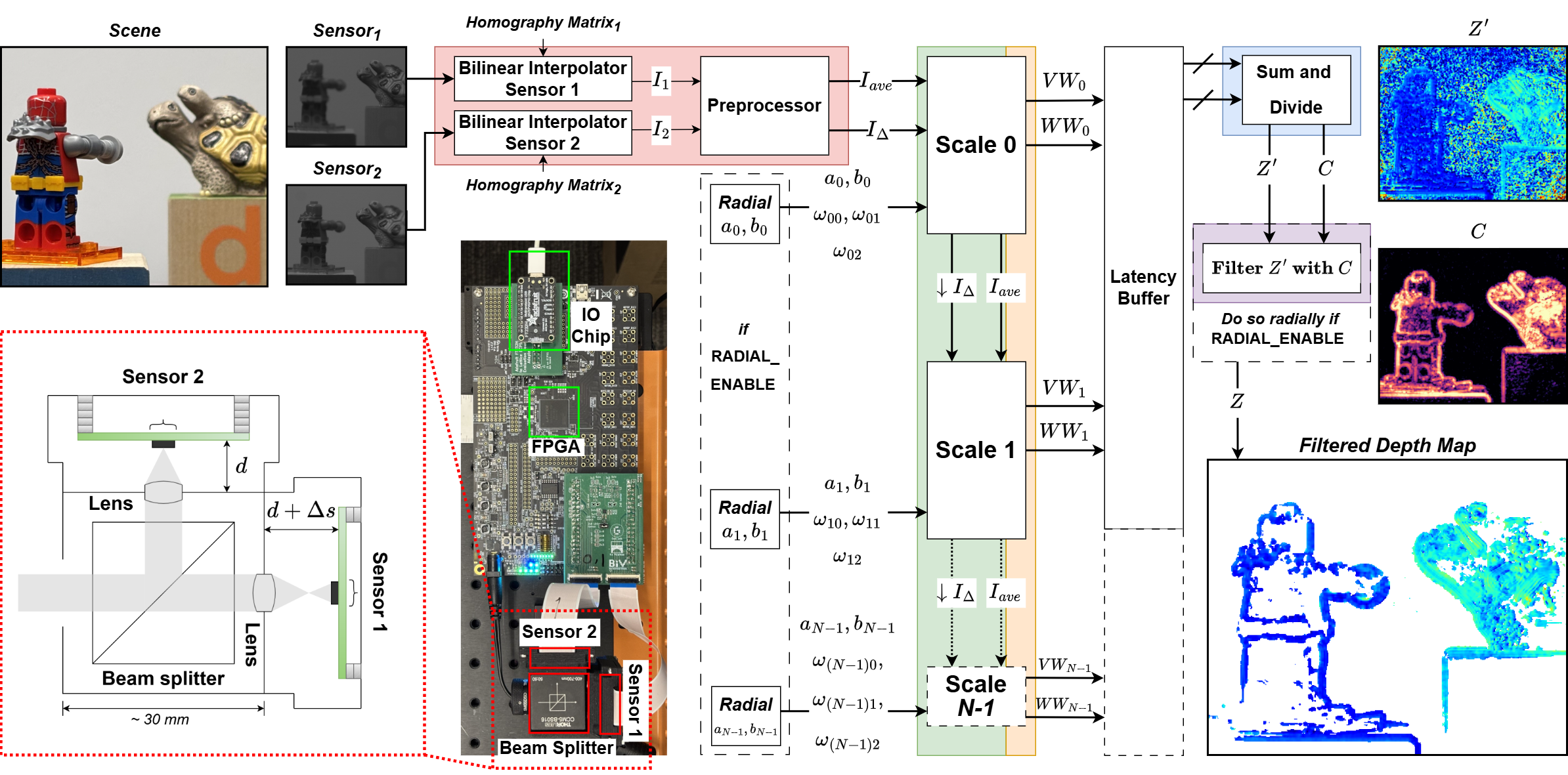}
    \caption{\textbf{Method overview.} Our camera uses a beam splitter and a pair of differentially-defocused low power sensors to observe the same scene with an offset depth of field. Our algorithm processes these images to produce depth maps thresholded by confidence.}
    \label{fig:method}
\end{figure*}

\subsection{Depth from Stereo}
Stereo vision is a passive depth imaging technique which uses two simultaneously captured images of the same scene with two different cameras spaced some distance apart~\cite{barnard1982computational}. By identifying translational differences in scene features between the two images. 
Stereo depth imaging systems are primarily concerned with identifying stereo matches~\cite{hamzah2016literature}. Stereo matching is extensively researched, with both classical and deep-learning methods attracting significant attention~\cite{hamzah2016literature,laga2020survey,poggi2021synergies}. Although deep-learning based stereo vision systems have achieved good results~\cite{zhou2020review,poggi2021synergies}, their significant compute and memory requirements rule out their appropriateness for space and power constrained systems. ~\cite{zhou2020review} compares 12 deep learning stereo vision methods, all of which run on Nvidia graphics cards (consuming upwards of 100 Watts~\cite{nvidiatitanx}) and most of which run at only a few frames per second.

\textbf{Hardware accelerators for stereo matching.} Classical methods for stereo matching involve evaluating the cost of different pixel matches - a patch of pixels associated with a region is compared with patches in other regions and a minimal-cost matching is determined for each pixel in the image. While searching the entire image typically gives the best stereo matching results, to be computationally tractable, most methods only perform a restricted local search~\cite{hamzah2016literature}. To make stereo depth cameras as compact and low-power as possible, significant work has been done in designing digital hardware to accelerate local stereo matching using FPGAs or ASICs.

Most hardware stereo matching accelerators work off of the same fundamental principle - two images are streamed in at the same time from two image sensors. Part of the image is stored on-chip in the FPGA and a census transform is used to implement semi-global matching~\cite{hirschmuller2007stereo,lu2021resource}. Ttofis \etal~pre-filters images using edge filters to make the matching task easier~\cite{ttofis2015low}. Zhang \etal reduces hardware overhead by using alternate transformations for cost calculation~\cite{zhang2018nipm}. Puglia \etal uses a dynamic programming inspired algorithm to adapt the problem to a systolic array, allowing for an efficient solution~\cite{puglia2017real}. While most of these works evaluate their implementations entirely in simulation on existing datasets, Mattoccia \etal demonstrated the real-world efficacy of these techniques by implementing a working, compact stereo depth camera~\cite{mattoccia2015passive}. Example depth maps from their device appear dense and accurate, but no quantitative metrics are reported so the working range cannot be evaluated. 

\subsection{Depth from Defocus}

Depth from defocus (DfD) estimates scene depth by analyzing image defocus~\cite{pentland1987new}. Traditional DfD approaches rely on computationally intensive convolution and deconvolution operations, resulting in high computational complexity~\cite{watanabe1998rational, levin2007image, zhou2011coded}.
Raj \& Staunton~\cite{joseph2014video} implement an efficient rational-filter-based DfD algorithm~\cite{raj2012rational} on an FPGA, reporting high real-world depth accuracy (0.5\% error) but over a very small working range (77-80 cm from the camera). Their system power is reported as 2 W, but this does not include sensor power or account for sequential rather than snapshot capture.
 
Our method falls into the DfDD class of algorithms~\cite{alexander2016focal, guo2017focal, alexander2018focal, guo2019compact, luo2025focal, luo2025depth,alexander2021depth}. Although DfDD has been shown to require dramatically fewer FLOPs than rational filter methods~\cite{luo2025depth}, its potential for low-power hardware implementation remains unexplored. To date, the lowest reported power consumption for a real-time DfDD sensor is still approximately 5~W~\cite{luo2025focal}.

\section{Method}
\label{sec:method}

Our camera uses a beam splitter and two low-power sensors to create a differential defocus signal that is processed into a depth map on a low-power FPGA. Similar to the optical design of Focal Split~\cite{luo2025focal}, one sensor is placed slightly farther from the beam splitter (see \cref{fig:method}), resulting in an offset depth of field. Typical of differential defocus methods ~\cite{alexander2019theory,guo2017focal,guo2019compact,luo2025focal,alexander2016focal,alexander2018focal,alexander2021depth}, comparisons $I_\Delta$ of brightness at corresponding pixels can be compared to spatial variations $\nabla^2I$ to cheaply reveal scene depth $Z$ at each pixel. Here, we simultaneously collect differentially defocused images $I_1$ and $I_2$, which are subtracted for $I_\Delta$ and averaged before the spatial derivatives in $\nabla^2I$. Then,
\begin{equation}
    \begin{aligned}
        Z(x,y) &= V(x,y)/W(x,y), \\
        V(x,y) &= a \nabla^2\tilde{I}(x,y), \\
        W(x,y) &= bV(x,y) - \tilde{I}_\Delta(x,y) ,
    \end{aligned}
    \label{eq:Z}
\end{equation}
where $a$ and $b$ are calibrated camera parameters discussed in \cref{sec:spatialvar} and $\Tilde{I}$ can refer to the original image, its spatial derivatives, or downsampled versions of the same~\cite{guo2017focal}. In natural scenes, depth cues are sparse but their presence and quality are visible in image data, allowing us to compute confidence for each depth estimate. In practice, several $\Tilde{I}$ variations are used to densify and robustify the output depth map, with a joint depth estimate thresholded by a joint confidence metric for the final depth map.

We introduce novel improvements to the camera and algorithm design before describing an efficient hardware implementation that enables 1.8$\times$ power savings over SOTA for the computation alone and 3.3$\times$ power savings over SOTA when measuring power consumed by the entire system operating on real hardware.

\subsection{Camera design}
\label{sec:camera}

Our overall camera design follows Focal Split~\cite{luo2025focal}, with a few changes to accommodate the smaller size of the low-power sensors (HM0360~\cite{hm0360manual}, which have a 12$\times$ power reduction vs. sensors in \cite{luo2025focal}). As shown in the system photograph in \cref{fig:method}, a 3D-printed enclosure connects each low power sensor to a lens on the beam splitter. In contrast to \cite{luo2025focal}, each sensor has its own lens, resulting in a FoV of 9.4$^{\circ}$ horizontal $\times$ 7.9$^{\circ}$ vertical (vs. 5.8$^{\circ} \times 4.3^{\circ}$ for \cite{luo2025depth}). A 10 mm achromatic lens was selected to strike a balance between a decent FoV and tolerable optical aberration. We provide a detailed parts list and DIY guide for assembling and calibrating our prototype in the supplement.

\subsection{Algorithm}
\label{sec:algorithm}

\definecolor{darkgreen}{rgb}{0.0,0.80,0.0}
\definecolor{violet}{rgb}{0.6, 0, 1}

Our algorithm, outlined in \cref{fig:method} and \cref{alg:spidercam}, processes the images in several stages. First, a {\sethlcolor{red!30}\hl{preprocessor}} applies a subpixel-accurate homography to correct for slight misalignments expected in camera assembly, followed by optional denoising by spatial filtering. The images are summed for the average image $I_{ave}$ and differenced for the change image $I_\Delta$. For each of two {\sethlcolor{green!30}\hl{image scales}}, $V$ and $W$ of~\cref{eq:Z} are computed for $\Tilde{I} \in \{I, I_x,I_y\}$. Each $\Tilde{I}$ contributes a separate hypothesis $(V_i,W_i)$ in parallel, which are {\sethlcolor{orange!30}\hl{partially summed}} before the latency buffer aligns the pixel streams across scales. The {\sethlcolor{blue!30}\hl{joint depth and confidence computation}} solves for depth and confidence from all 6 estimates at each pixel with a novel method described below. Finally,  a novel {\sethlcolor{violet!30}\hl{spatially-varying confidence threshold is applied}}, producing the output depth map.

\begin{algorithm}
    \caption{\textbf{The SpiderCam Algorithm.} This DfDD algorithm achieves SOTA power efficiency with memory locality, effective use of efficient kernels, and robustness to the elevated noise levels of low-power sensors.}
    \label{alg:spidercam}
    \begin{algorithmic}[1] 
        \STATE Input: images $I_{1}$, $I_{2}$, calibrated parameters $(\{a_0,...,a_{N-1}\},\{b_0,...,b_{N-1}\}, \{\omega_0, ..., \omega_{(3N - 1)}\}$, $C_{thresh}, Z_{min}, Z_{max})$
        \STATE Output: depth map $Z(x,y)$
        
        \color{red}
        \STATE $I_{ave} \gets \texttt{denoise}(I_{1} + \texttt{homography}(I_{2}))/2$
        \STATE $I_\Delta \gets \texttt{denoise}(I_{1} - \texttt{homography}(I_{2}))/2$
        
        \color{darkgreen}
        \STATE $V, W, VW, WW \gets [~~]$
        \STATE $VW_+, WW_+ \gets [~~]$
        
        \FOR {$scale$ in $\{0,1,...,N-1\}$}
             
            \STATE $I_{lap} \gets \texttt{laplacian}(I_{ave})$
            \FOR {$d$ in \{$none$, $\partial_x$, $\partial_y\}$}
                \STATE
                $V$.push$(a_{scale} \times \texttt{deriv}(I_{lap}, d))$
                \STATE 
                $W$.push$(b_{scale} \times V[-1] - \texttt{deriv}(I_\Delta, d))$
            \ENDFOR

            \FOR {$i$ in \{$scale\times 3$, $scale\times 3 + 2$\}}
                \STATE $ VW$.push$(\omega_{i} \times  V[i] \times W[i])$
                \STATE $ WW$.push$(\omega_{i} \times  W[i] \times W[i])$
                \STATE $ VW[i] \gets \texttt{upsample}( VW[i], scale)$
                \STATE $ WW[i] \gets \texttt{upsample}( WW[i], scale)$
            \ENDFOR
            
            \STATE $I_{ave} \gets \texttt{downsample}(I_{ave}, 1)$
            \STATE $I_{\Delta} \gets \texttt{downsample}(I_\Delta, 1)$
    
    \color{orange}
            \STATE $VW_{+}$.push($\texttt{sum}(VW[scale\times3:scale\times3+2])$)
            \STATE $WW_{+}$.push($\texttt{sum}(WW[scale\times3:scale\times3+2])$)
        \color{green}
        \ENDFOR

        \color{blue}  

        \STATE $C \gets \texttt{sum}(VW_+)$
        \STATE $Z \gets \texttt{divide} (C, \texttt{sum}(WW_+))$
        
        \color{violet}
        
        \STATE $Z[C<C_{thresh} \text{~or not~} Z_{min}<Z<Z_{max}] \gets \texttt{null}$
    \end{algorithmic}
\end{algorithm}

\subsection{Joint Depth Estimation}
Previous DfDD methods~\cite{guo2017focal,guo2019compact} have combined multiple per-pixel depth estimates $\{Z_i(x,y)\}$ with softmax weighted by a per-pixel, per-estimate, dynamically computed confidence $\{C_i(x,y)\}$:
\begin{align}
    Z(x,y) = \frac{\sum_{i} e^{C_i(x,y)} Z_i(x,y)}{\sum_{i} e^{C_i(x,y)}}.
\end{align}
While successful in densifying and robustifying output depth, this expression is impossible to compute under our resource constraints. Efficient softmax hardware implementation is an active area of research; without cutting-edge optimizations, the softmax operation would consume more FPGA resources than the entirety of some configurations of our system~\cite{hu2018efficient,geng2018hardware}. Instead, we compute a joint depth estimate from 6 estimates $i$ (based on 2 scales $\times$ 3 derivative orders) using calibrated weights $\omega_i$ as
\begin{align}
    Z(x,y) = \frac{\sum_i \omega_i V_i(x,y) W_i(x,y)}{\sum_i \omega_i W_i(x,y)^2}.
    \label{eq:jointZ}
\end{align}
Note that~\cref{eq:jointZ} requires only a single division, which is important because division costs as much as 10 additions in our setting. Improvements to the working range due to this joint estimation can be seen by comparing to the green curve in \cref{fig:quant_analysis}a and the small blue icons in \cref{fig:tradeoffs}.

\subsection{Confidence Estimation}
Because triangulation cues are sparse in natural scenes, a per-pixel confidence score is needed to exclude non-informative image regions where \cref{eq:Z} will provide incorrect output. Previous DfDD methods have used more expensive~\cite{guo2017focal,guo2019compact} or less accurate~\cite{luo2025depth,luo2025focal} confidence metrics. We show that a cheap and effective confidence combining across estimates $i$ at each pixel is: 
\begin{align}
    C_i(x,y) =&  V_i(x,y)W_i(x,y), \\
    C(x,y) =& \sum_i  \omega_i C_i(x,y). \label{eq:C}
\end{align}
This simple sum enables us to robustify and densify our measurements across estimates as in~\cite{guo2017focal,guo2019compact} without the added expense of their more complex per-pixel estimates or the softmax-based combination previously used. In comparison to the estimator used in~\cite{luo2025depth,luo2025focal}, we report an increase in working range of 5cm (See~\cref{fig:conf}).
We note that as the numerator of our depth estimation, calculating $C$ requires no additional computation. Low confidence regions will occur when texture is unavailable (so $V \propto \nabla^2 I$  is close to zero) and unstable divisions (when $W$ and therefore the denominator $W^2$ is close to zero). 

\subsection{Calibration}
\label{sec:calibration}
Our depth computation requires a handful of calibrated parameters, specifically $a$ and $b$ from \cref{eq:Z}, $\{\omega_i\}$ in \cref{eq:jointZ} and \cref{eq:C}, and thresholds $C_{thresh}, Z_{min}, Z_{max}$ applied to the output. Each depth estimate uses $a$ and $b$, which physically correspond to
\begin{align}
    a = -A^2, ~~~b =\left(\rho - 1/s\right),
\end{align}
for aperture diameter $A$, lens optical power $\rho$, and lens-to-sensor distance $s$~\cite{luo2025focal}. While these parameters could be derived directly camera measurements, we instead optimize them for depth performance. Calibration of $\{\omega_i\}$ is also data-driven. We collect a real-world 
calibration dataset by placing a plane with printed texture at 56 known depths, evenly spaced from 0.24m to 1.36m, and optimize the parameters using mean absolute depth error as the loss function. To improve accuracy, we only compute the loss on the 10\% most confident pixels in each radial region for each image pair to prevent overfitting to noise. Finally, we establish confidence thresholds for each radial zone to maximize the working range, see \cref{sec:suppcalibration} for details.

\begin{figure*}[ht!]
    \centering
    \includegraphics[width=1\linewidth]{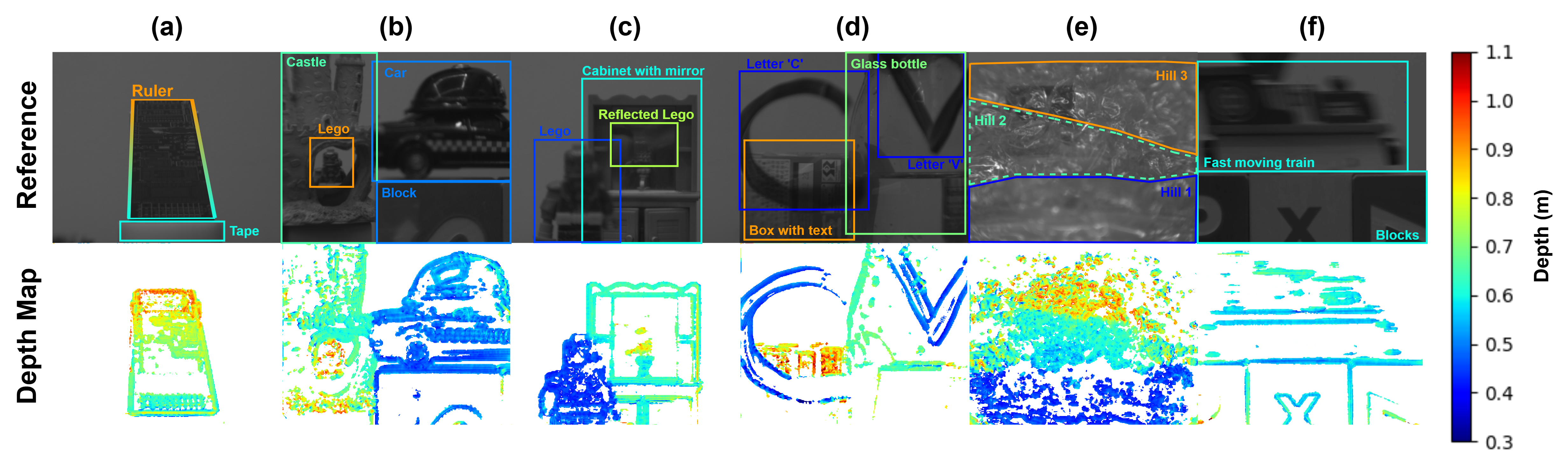}
    \caption{\textbf{Captured images and depth maps computed on-device.} Note the handling of continuous depths (vs. disparity levels), reflective textures, and transparent objects that active sensing would struggle with, as well as objects in motion due to snapshot capture.}
    \label{fig:real_scenes}
\end{figure*}

\subsection{Spatial Variation}
\label{sec:spatialvar}
The optics available for the compact size of our low-power sensors create noticeable spatial variation in defocus due to Petzval field curvature~\cite{hecht2017optics} and other nonidealities. Because of this, performance suffers if we hold our $a$, $b$, and thresholds to constant values across the image sensor. We improve performance by allowing calibrated parameters to vary for 16 separate radial zones of equal width. Since the pixels arrive in order in the pixel stream, we are able to track the radial locations of each pixel without having to actually buffer the (x,y) information alongside each pixel value. By comparing a streaming value to thresholds on squared radial distances (to avoid the expensive square root operation), we can dynamically update the values of $(\{a_0,...,a_{N-1}\},\{b_0,...,b_{N-1}\}, C_{thresh}, Z_{min}, Z_{max})$, see in detail \cref{subsec:rz_adapt_supp}. Note that $\omega_i$ values do not vary spatially.
Accounting for spatial variation results in a larger working range (\cref{fig:quant_analysis}a) and better calibrated depth for each radial region (\cref{fig:heatmap_rings}) with only a small increase in core power (\cref{fig:tradeoffs}).

\subsection{Efficient Hardware Implementation}
FPGA size and resource utilization strongly affects system power consumption - larger FPGAs have higher static power dissipation. In order to make our system as efficient as possible, we designed our hardware to fit inside the Lattice ECP5 LFE5U-85F FPGA. By using a smaller FPGA, the benefits of our efficient algorithm implementation are realized in power savings.

We also had to restrict our memory use to achieve low-power operation. Low-power FPGAs have extremely limited memory budgets. Due to the high energy cost of data movement, extending their memory by using off-chip DRAM incurs a prohibitively high energy cost~\cite{ghose2018yourdram}. The LFE5U-85F does not have enough on-chip memory to hold both frames at once. Adding enough storage to buffer both frames without moving to a larger, more power-hungry FPGA would require the use of external DRAM, raising power consumption and incurring depth map latency. This means that we must process our depth maps with a data streaming model, where the data locality of the DfDD algorithm enables significant power savings. Additionally, we developed a hybrid fixed and floating point architecture that supports the high dynamic range of natural scenes while minimizing per-pixel memory cost.

\subsubsection{Multiscale compute within fixed memory budget}
Our robust depth algorithm requires upsampling in two of its computations: estimating $\nabla^2\tilde{I}$ with a Laplacian pyramid, and merging downsampled estimates into our multiscale joint estimation. 

Downsampling a streaming image is straightforward, but naive upsampling requires buffering a prohibitive amount of the image. We develop a specialized streaming upsampler using zero-interleaved kernels that reach a larger effective receptive field without increasing arithmetic operations. This both decreases the number of lines buffered and fixes the memory use to be independent of image height. See \cref{sec:upsampling_supp} for details. 

\subsubsection{Efficient Kernels}
We minimize FPGA power and memory costs by using kernels that are small, have integer coefficients that are powers of two, and are linearly separable. This, respectively, minimizes the number of lines buffered, converts multiplication to bit-shifting, and reduces convolution costs by a factor of kernel width. We use box, Gaussian, and derivative kernels in the preprocessing and per-scale computations, with our largest kernel being a 5-tap Burt-Adelson Gaussian filter~\cite{burt1987laplacian}. See \cref{sec:efficient_kernels} for details.

\subsubsection{Hybrid Fixed Point with Efficient Floating Point}
Fixed point arithmetic is cheap in hardware, but limits the dynamic range of values that can be represented accurately. Most vision applications require floating point, which is significantly more expensive. We reduce this cost by identifying preprocessing steps that can be computed in fixed point without loss of accuracy, then switching to an efficient floating point implementation that maximizes accuracy within a constrained memory footprint.

Specifically, the homography and preprocessing stages are implemented in fixed point. The homography only requires a small number of lines to be buffered since the affine transformation is small. In the preprocessing step, our carefully selected filters do not require bit trimming, avoiding quantization issues. Due to the high dynamic range of our images, after these stages, the results are converted to floating point before computing $I_{ave}$ and $I_{\Delta}$.

On the floating-point side, we save FPGA resources by dropping subnormal number support and using FP16 instead of FP32. Removing subnormal number support makes multipliers and dividers nearly as inexpensive as fixed-point arithmetic, with only a small additional cost for exponent handling. Our adders still require the variable shifters and priority encoders associated with floating-point arithmetic, but some of the surrounding logic is simplified by removing subnormal handling. 
The primary reduction in resource usage for our floating-point adders comes from using half-precision (FP16) representation. See ~\cref{sec:hf_fp_supp} for details.

\subsubsection{Data Readout}
Our system outputs 8-bit depth maps over a parallel port, similar to how an image sensor outputs its data. To read the results of the depth computation from the camera, we use an FTDI FT232H parallel-to-USB interface chip. Our communication speed is limited to 8MB/s by the design of the ECP5 development board that our prototype system uses, meaning that in-situ our system can output depth maps at a maximum of 32.5 FPS. Alternatively, for data collection and demonstration, we can read out image pairs or one image with its respective depth map at a slower rate. Without these I/O limitations, our core computation could run up to 140 FPS on ECP5 and up to 300 FPS on Kintex-7.

\begin{figure*}[!t]
    \centering
    \includegraphics[width=1\linewidth]{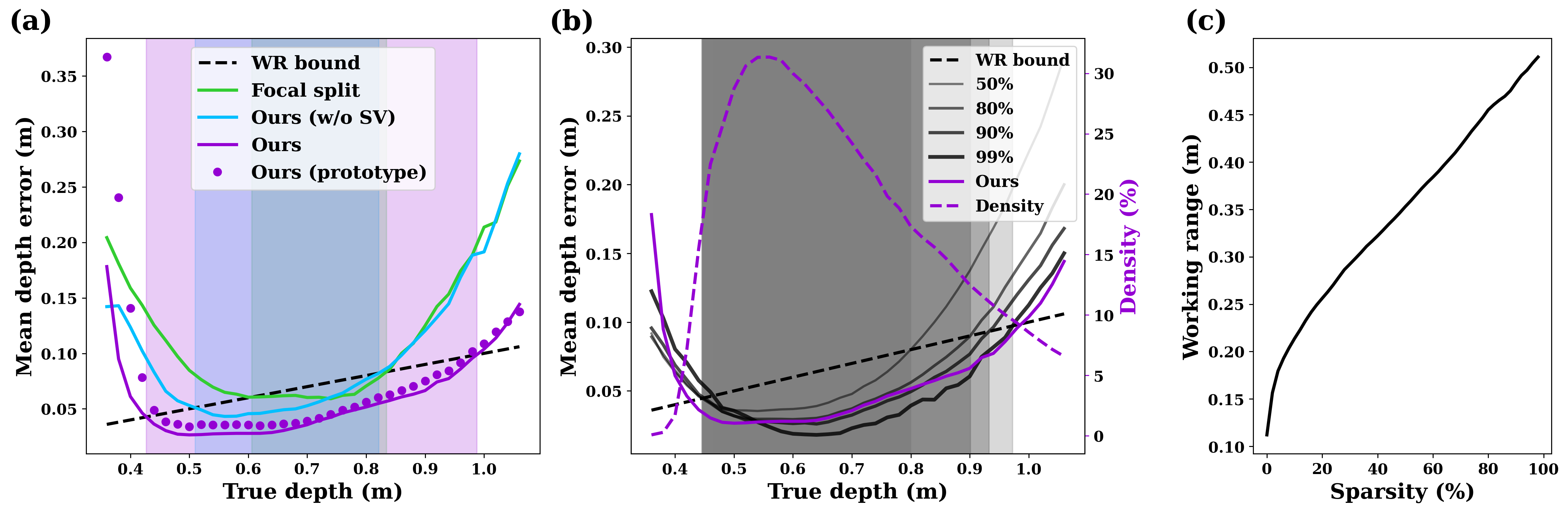}
    \caption{
        {\bf Quantitative analysis on images from device.}
        (a) Mean absolute error (MAE) of depth as a function of true depth, with working range bounds.
        Our method robustly outperforms the Focal Split algorithm~\cite{luo2025focal} in the low power hardware setting, and accounting for spatial variation in our compact optics is vital. 
        Purple dots show errors from depth maps computed on-device and demonstrate good agreement with off-line depth computations (rest of figure), which were performed in 32-bit floating point.
        (b) The MAE vs. depth curve of our method with dynamic sparsity values set by per-image confidence percentiles (grays) and the MAE (solid purple) and output pixel density (100-sparsity, dashed purple) resulting from the fixed confidence thresholds used in our method.
        (c) Working range increases as data is sparsified by per-image confidence percentiles.
        }
    \label{fig:quant_analysis}
\end{figure*}

\begin{figure}[h]
    \centering
    \includegraphics[width=1\linewidth]{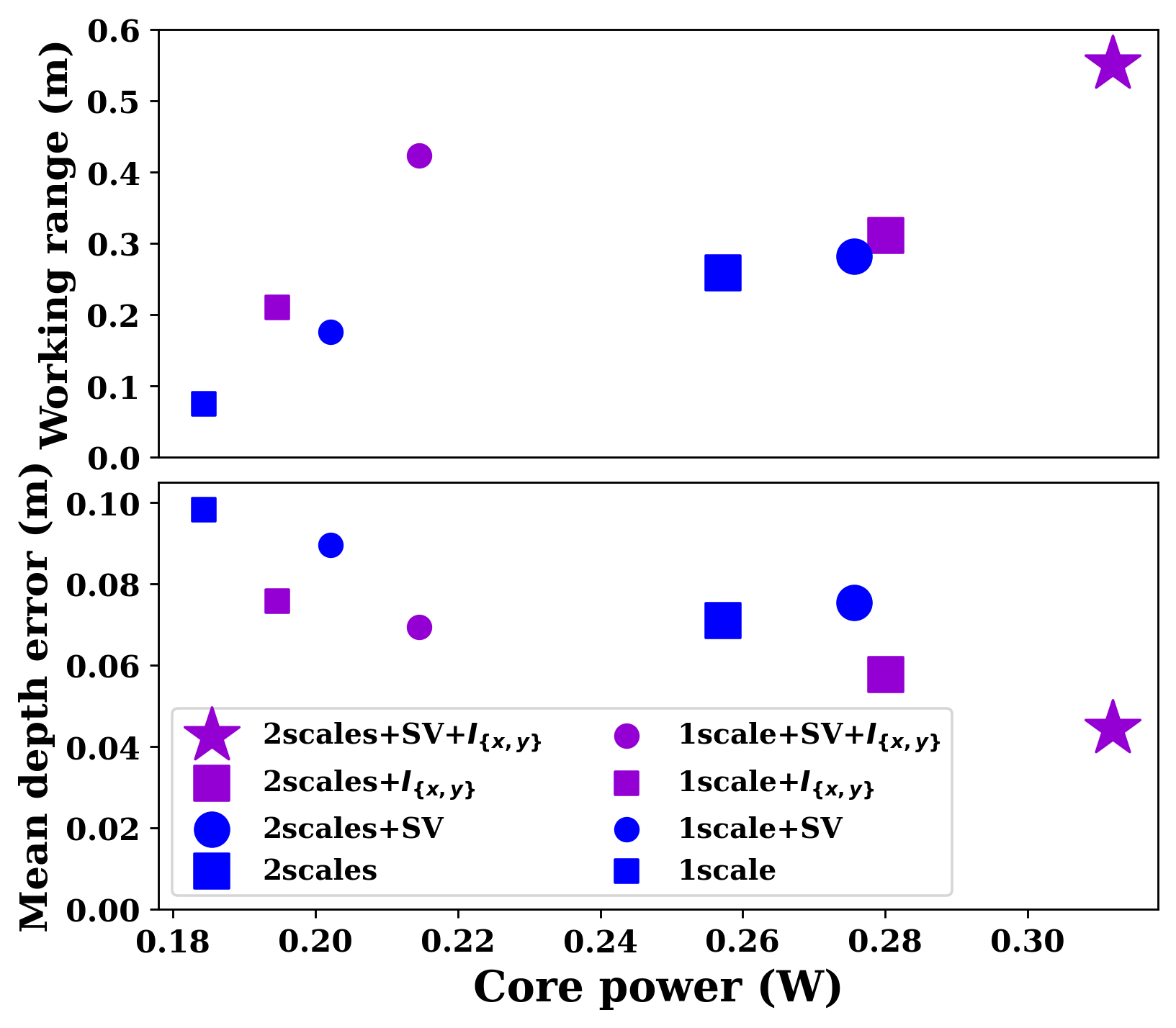}
    \caption{\textbf{Power/accuracy trade-offs for alternate algorithms.} We estimate max core power on the ECP5 with manufacturer's tools and evaluate working range and MAE from 0.4m to 1.0m offline on images collected by our hardware prototype. Adding computational components improves accuracy while increasing power cost. Our algorithm is shown with the purple star. 
    }
    \label{fig:tradeoffs}
\end{figure}

\begin{figure*}[!t]
    \centering
    \includegraphics[width=1\linewidth]{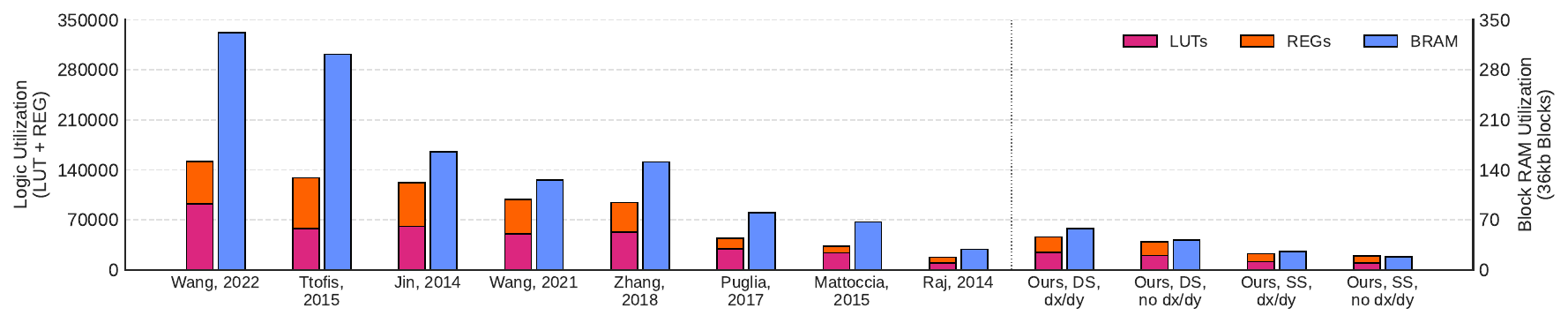}
    \caption{\textbf{Resource Usage Across Methods.} 
    We compare the resource utilization of several configurations of our system as well as systems in the literature. ``DS" and ``SS" denote evaluation at 2 and 1 scales, respectively.}
    \label{fig:resource}
\end{figure*}

\section{Results}
\label{sec:results}

Depth maps of real scenes computed on-device are shown in \cref{fig:real_scenes}, along with corresponding sensor images. Our computed confidence excludes low-signal regions, returning accurate depth estimates around scene texture. Note that our method can gracefully handle scenes with multiple depths, including continuous (a), discontinuous (b), and reflected (c) depths. Our sparse depth maps gracefully handle overlapping transparent objects (d), our passive method easily handles bubble wrap (e) even though it is transparent, reflective, and non-rigidly deformed, and our snapshot camera returns accurate depth even in the presence of significant motion blur (f). Additional scenes and videos are provided in the supplement.

We characterize the performance of the system using front-parallel textured planes across densely-sampled depths, and define the working range (WR) as the set of depths for which $|Z_{est}-Z_{true}|<0.1 ~Z_{true}$, with our device achieving a working range of 0.45-0.97 m. Accuracy of on-device computation (purple dots) shows good agreement with off-device computation (purple and all other lines) in \cref{fig:quant_analysis}a. Off-device computation is slightly more accurate because it is performed fully in 32-bit floating point for convenience and fairness in comparing across algorithms.
 
In comparison to our method (offline) with a working range (pale purple highlight) of 0.43-0.99 m, the Focal Split algorithm~\cite{luo2025focal} (green) has a working range of only 0.60-0.83 m on the challenging low-power camera data. Spatial variation is a major contributor to this improvement, as removing it (blue) drops our working range to 0.51-0.82 m.

Next, to explore the effect of confidence on sparsity and accuracy,~\cref{fig:quant_analysis}b shows the MAE and WR for depth- and scene-varying confidence thresholds set to enforce a fixed sparsity across each depth map (gray levels). Our method, repeated in purple for reference, takes a fixed confidence threshold that naturally loses density (100\% - sparsity, dashed purple) at the edge of the WR. Note that fixed thresholds result in different error/density trade-offs than enforced sparsities, highlighting the need for computationally inexpensive, well-calibrated confidence estimation as an area for future research. \cref{fig:quant_analysis}c shows the size of the working range across enforced sparsities.

\subsection{Resource Utilization and Power Consumption}
\textbf{DfDD core power and size.} To evaluate the efficiency of the SystemVerilog DfDD implementation itself, we synthesized multiple configurations of the DfDD core depth computation. 
Even though our physical system is implemented with a Lattice ECP5 FPGA, for the sake of this specific resource usage comparison, we synthesized our design for the more widely-used Xilinx 7-Series FPGA architecture, which is widely compatible across the literature. Resource utilization results can be seen in~\cref{fig:resource}, details in \cref{sec:power_estimates_supp}. Each FPGA has a limited number of lookup tables (LUT), registers (REG), and block RAM (BRAM) with which a circuit design can be implemented. A smaller design consumes less resources and can fit on smaller FPGAs, allowing for lower power consumption and better system miniaturization. We also took these utilization numbers and used vendor tools to estimate power consumption, comparing our system's power consumption with similar work~\cref{tab:powertable}.

\textbf{Power consumption of hardware camera.} Critical to our work is the construction of an actual hardware camera capable of producing depth images in real-time at over 30 FPS. Our hardware camera consisted of two HM0360 low-power image sensors in a custom optical setup (as described in~\cref{sec:camera}) connected via a custom interposer PCB to a commercially available Lattice ECP5 evaluation board~\cite{ecp5evalmanual} equipped with a LFE5UM5G-85F FPGA~\cite{ecp5familydatasheet}. In order to reduce power consumption, some modifications were made to the evaluation board. A summary of these modifications can be found in~\cref{sec:diy_electrical_supp}.
To measure power consumption, we powered the camera entirely by an Otii Arc power profiler and operated it under normal conditions, recording natural scenes, integrating average power consumption over 5 seconds. Our results are shown across configurations in~\cref{tab:wallpower}. Significantly, our full algorithm consumes only 624 mW at 32.5 FPS and 399 mW at 9 FPS, 3.3 - 5$\times$ less power than the lowest full system power reported in the literature ~\cite{puglia2017real}, which runs at 30 FPS.

Total system power measurements in~\cref{tab:wallpower} are higher than core power estimates in~\cref{tab:powertable,fig:ablations} because the physical hardware system 
also includes: two image sensors and image homography and cropping logic, I/O circuitry, and the evaluation board which contains several unused components and introduces inefficiencies in, e.g., voltage converters.

\subsection{Accuracy Across Restricted Power Budgets}
\label{sec:ablation}

We analyze ablations of several subcomponents in terms of both depth performance, evaluated by working range and depth MAE over 0.4 m to 1.0 m (computed off-device), and core power consumption (estimated for the ECP5 from manufacturer's tools). \cref{fig:tradeoffs} provides an overview of these trade-offs.  
Adding spatial variation in calibrated parameters improves working range and reduces MAE (see corresponding square-circle pairs). As expected, the accuracy among well-calibrated (i.e. spatially varying) methods goes according to the number of separate estimates used in the depth map: the best has 6 (purple star: 2 scales, 3 derivatives), the next best has 3 (small purple circle: 1 scale, 3 derivatives), the next best has 2 (large blur circle: 2 scales, 1 derivative), and the worst has only 1 (small blue circle: 1 scale, 1 derivative). Adding derivatives is cheaper in terms of power consumption than adding image scales, though we note that the relative cost appears more extreme in estimates of core power than the true relative difference in the overall system, since peripheral power costs dominate and will be constant across methods. Specifically, in comparing our method with the next best option (single scale, multiderivative, shown by the purple circle), we estimate a core power difference of 311 vs. 214 mW for a roughly 32\% difference, but when running the hardware prototype at 32.5 FPS we measure the full systems' wall power at 624 vs. 489 mW -- only a 20\% difference (dropping to an 11\% difference at 9 FPS, see~\cref{tab:wallpower}). At a high level, this study indicates the potential of our method to provide depth estimates for a range of power constraints through appropriate computational trade-offs.

\section{Conclusion}
\label{sec:discussions}

We present SpiderCam, the first sub-Watt FPGA-based passive 3D camera. Featuring snapshot capture with compact off-the-shelf optics, low-power sensors, a robust and efficient new DfDD algorithm, and a SystemVerilog implementation on a low-power FPGA, we demonstrate a working range of 52 cm at 32.5 FPS in real time on real data with a full-system wall power of 624 mW.
We have characterized the power/accuracy trade-offs of several variations of our system, and identified improvements in optics and confidence estimation as immediate future directions for system improvement. Significant gains in power consumption could be pursued by moving from an FPGA to an ASIC, which typically results in 5-20$\times$ power savings depending on design details~\cite{kuon2006measuring,mamish2024nir}. 
Our sparse and somewhat noisy depth estimates offer exciting possibilities as control signals for embedded/edge applications like wearables and robotics, which will require further research and development. An immediate direction for future work is the integration of our outputs with efficient depth-completion pipelines like~\cite{zuo2025omni} and increasing the FoV of our system.

\section*{Acknowledgments}
This research was partially supported by the National Science Foundation under awards numbers CNS-2430327 and CCF-2431505.
Any opinions, findings, conclusions, or recommendations expressed in this material are those of the authors and do not necessarily reflect the views of the National
Science Foundation or other supporters.
We would also like to thank Junjie Luo and Alan Fu for helpful discussions.
{
    \small
    \bibliographystyle{ieeenat_fullname}
    \bibliography{main, sec/power_comparison}

@String(CVPR= {IEEE Conf. Comput. Vis. Pattern Recog.})

@String(TOG= {ACM Trans. Graph.})

@String(CVPR  = {CVPR})

@String(TOG   = {ACM TOG})

@inproceedings{Geiger2012CVPR,
  author = {Andreas Geiger and Philip Lenz and Raquel Urtasun},
  title = {Are we ready for Autonomous Driving? The KITTI Vision Benchmark Suite},
  booktitle = {Conference on Computer Vision and Pattern Recognition (CVPR)},
  year = {2012}
}

@article{Geiger2013IJRR,
  author = {Andreas Geiger and Philip Lenz and Christoph Stiller and Raquel Urtasun},
  title = {Vision meets Robotics: The KITTI Dataset},
  journal = {International Journal of Robotics Research (IJRR)},
  year = {2013}
}

@inproceedings{Fritsch2013ITSC,
  author = {Jannik Fritsch and Tobias Kuehnl and Andreas Geiger},
  title = {A New Performance Measure and Evaluation Benchmark for Road Detection Algorithms},
  booktitle = {International Conference on Intelligent Transportation Systems (ITSC)},
  year = {2013}
}

@inproceedings{Menze2015CVPR,
  author = {Moritz Menze and Andreas Geiger},
  title = {Object Scene Flow for Autonomous Vehicles},
  booktitle = {Conference on Computer Vision and Pattern Recognition (CVPR)},
  year = {2015}
}

@article{scharstein2002taxonomy,
  title={A taxonomy and evaluation of dense two-frame stereo correspondence algorithms},
  author={Scharstein, Daniel and Szeliski, Richard},
  journal={International journal of computer vision},
  volume={47},
  number={1},
  pages={7--42},
  year={2002},
  publisher={Springer}
}

@inproceedings{scharstein2003high,
  title={High-accuracy stereo depth maps using structured light},
  author={Scharstein, Daniel and Szeliski, Richard},
  booktitle={2003 IEEE Computer Society Conference on Computer Vision and Pattern Recognition, 2003. Proceedings.},
  volume={1},
  pages={I--I},
  year={2003},
  organization={IEEE}
}

@inproceedings{scharstein2007learning,
  title={Learning conditional random fields for stereo},
  author={Scharstein, Daniel and Pal, Chris},
  booktitle={2007 IEEE conference on computer vision and pattern recognition},
  pages={1--8},
  year={2007},
  organization={IEEE}
}

@inproceedings{hirschmuller2007evaluation,
  title={Evaluation of cost functions for stereo matching},
  author={Hirschmuller, Heiko and Scharstein, Daniel},
  booktitle={2007 IEEE conference on computer vision and pattern recognition},
  pages={1--8},
  year={2007},
  organization={IEEE}
}

@inproceedings{scharstein2014high,
  title={High-resolution stereo datasets with subpixel-accurate ground truth},
  author={Scharstein, Daniel and Hirschm{\"u}ller, Heiko and Kitajima, York and Krathwohl, Greg and Ne{\v{s}}i{\'c}, Nera and Wang, Xi and Westling, Porter},
  booktitle={German conference on pattern recognition},
  pages={31--42},
  year={2014},
  organization={Springer}
}

@article{land1969structure,
  title={Structure of the retinae of the principal eyes of jumping spiders (Salticidae: Dendryphantinae) in relation to visual optics},
  author={Land, MF},
  journal={Journal of experimental biology},
  volume={51},
  number={2},
  pages={443--470},
  year={1969},
  publisher={The Company of Biologists Ltd}
}

@article{nagata2012depth,
  title={Depth perception from image defocus in a jumping spider},
  author={Nagata, Takashi and Koyanagi, Mitsumasa and Tsukamoto, Hisao and Saeki, Shinjiro and Isono, Kunio and Shichida, Yoshinori and Tokunaga, Fumio and Kinoshita, Michiyo and Arikawa, Kentaro and Terakita, Akihisa},
  journal={Science},
  volume={335},
  number={6067},
  pages={469--471},
  year={2012},
  publisher={American Association for the Advancement of Science}
}

@article{raj2012rational,
  title={Rational filter design for depth from defocus},
  author={Raj, Alex Noel Joseph and Staunton, Richard C},
  journal={Pattern Recognition},
  volume={45},
  number={1},
  pages={198--207},
  year={2012},
  publisher={Elsevier}
}

@book{giancola2018survey,
  title={A survey on 3D cameras: Metrological comparison of time-of-flight, structured-light and active stereoscopy technologies},
  author={Giancola, Silvio and Valenti, Matteo and Sala, Remo},
  year={2018},
  publisher={Springer}
}

@article{horaud2016overview,
  title={An overview of depth cameras and range scanners based on time-of-flight technologies},
  author={Horaud, Radu and Hansard, Miles and Evangelidis, Georgios and M{\'e}nier, Cl{\'e}ment},
  journal={Machine vision and applications},
  volume={27},
  number={7},
  pages={1005--1020},
  year={2016},
  publisher={Springer}
}

@article{behroozpour2017lidar,
  title={Lidar system architectures and circuits},
  author={Behroozpour, Behnam and Sandborn, Phillip AM and Wu, Ming C and Boser, Bernhard E},
  journal={IEEE Communications Magazine},
  volume={55},
  number={10},
  pages={135--142},
  year={2017},
  publisher={IEEE}
}

@article{geng2011structured,
  title={Structured-light 3D surface imaging: a tutorial},
  author={Geng, Jason},
  journal={Advances in optics and photonics},
  volume={3},
  number={2},
  pages={128--160},
  year={2011},
  publisher={Optical Society of America}
}

@article{raj2020survey,
  title={A survey on LiDAR scanning mechanisms},
  author={Raj, Thinal and Hanim Hashim, Fazida and Baseri Huddin, Aqilah and Ibrahim, Mohd Faisal and Hussain, Aini},
  journal={Electronics},
  volume={9},
  number={5},
  pages={741},
  year={2020},
  publisher={MDPI}
}

@article{lee2020accuracy,
  title={Accuracy--power controllable LiDAR sensor system with 3D object recognition for autonomous vehicle},
  author={Lee, Sanghoon and Lee, Dongkyu and Choi, Pyung and Park, Daejin},
  journal={Sensors},
  volume={20},
  number={19},
  pages={5706},
  year={2020},
  publisher={MDPI}
}

@manual{realsensemanual,
    organization  = "Intel",
    title         = "Intel RealSense D400 Series Product Family Datasheet",
    number        = "D400",
    year          =  2019,
    month         =  1
}

@article{bhoi2019monocular,
  title={Monocular depth estimation: A survey},
  author={Bhoi, Amlaan},
  journal={arXiv preprint arXiv:1901.09402},
  year={2019}
}

@article{ming2021deep,
  title={Deep learning for monocular depth estimation: A review},
  author={Ming, Yue and Meng, Xuyang and Fan, Chunxiao and Yu, Hui},
  journal={Neurocomputing},
  volume={438},
  pages={14--33},
  year={2021},
  publisher={Elsevier}
}

@article{masoumian2022monocular,
  title={Monocular depth estimation using deep learning: A review},
  author={Masoumian, Armin and Rashwan, Hatem A and Cristiano, Juli{\'a}n and Asif, M Salman and Puig, Domenec},
  journal={Sensors},
  volume={22},
  number={14},
  pages={5353},
  year={2022},
  publisher={MDPI}
}

@article{ozyecsil2017survey,
  title={A survey of structure from motion*.},
  author={{\"O}zye{\c{s}}il, Onur and Voroninski, Vladislav and Basri, Ronen and Singer, Amit},
  journal={Acta Numerica},
  volume={26},
  pages={305--364},
  year={2017},
  publisher={Cambridge University Press}
}

@inproceedings{schonberger2016structure,
  title={Structure-from-motion revisited},
  author={Schonberger, Johannes L and Frahm, Jan-Michael},
  booktitle={Proceedings of the IEEE conference on computer vision and pattern recognition},
  pages={4104--4113},
  year={2016}
}

@article{saputra2018visual,
  title={Visual SLAM and structure from motion in dynamic environments: A survey},
  author={Saputra, Muhamad Risqi U and Markham, Andrew and Trigoni, Niki},
  journal={ACM Computing Surveys (CSUR)},
  volume={51},
  number={2},
  pages={1--36},
  year={2018},
  publisher={ACM New York, NY, USA}
}

@article{laga2020survey,
  title={A survey on deep learning techniques for stereo-based depth estimation},
  author={Laga, Hamid and Jospin, Laurent Valentin and Boussaid, Farid and Bennamoun, Mohammed},
  journal={IEEE transactions on pattern analysis and machine intelligence},
  volume={44},
  number={4},
  pages={1738--1764},
  year={2020},
  publisher={IEEE}
}

@article{hamzah2016literature,
  title={Literature survey on stereo vision disparity map algorithms},
  author={Hamzah, Rostam Affendi and Ibrahim, Haidi},
  journal={Journal of Sensors},
  volume={2016},
  number={1},
  pages={8742920},
  year={2016},
  publisher={Wiley Online Library}
}

@article{barnard1982computational,
  title={Computational stereo},
  author={Barnard, Stephen T and Fischler, Martin A},
  journal={ACM Computing Surveys (CSUR)},
  volume={14},
  number={4},
  pages={553--572},
  year={1982},
  publisher={ACM New York, NY, USA}
}

@article{poggi2021synergies,
  title={On the synergies between machine learning and binocular stereo for depth estimation from images: A survey},
  author={Poggi, Matteo and Tosi, Fabio and Batsos, Konstantinos and Mordohai, Philippos and Mattoccia, Stefano},
  journal={IEEE Transactions on Pattern Analysis and Machine Intelligence},
  volume={44},
  number={9},
  pages={5314--5334},
  year={2021},
  publisher={IEEE}
}

@article{zhou2020review,
  title={Review of stereo matching algorithms based on deep learning},
  author={Zhou, Kun and Meng, Xiangxi and Cheng, Bo},
  journal={Computational intelligence and neuroscience},
  volume={2020},
  number={1},
  pages={8562323},
  year={2020},
  publisher={Wiley Online Library}
}

@online{nvidiatitanx,
  author = {MultiMedia LLC},
  title = {Product info - GeForce GTX TITAN X},
  year = 2025,
  url = {https://web.archive.org/web/20250828153625/https://www.nvidia.com/en-us/geforce/graphics-cards/geforce-gtx-titan-x/specifications/}
}

@article{hirschmuller2007stereo,
  title={Stereo processing by semiglobal matching and mutual information},
  author={Hirschmuller, Heiko},
  journal={IEEE Transactions on pattern analysis and machine intelligence},
  volume={30},
  number={2},
  pages={328--341},
  year={2007},
  publisher={IEEE}
}

@article{ngo2025edge,
  title={Edge Intelligence: A Review of Deep Neural Network Inference in Resource-Limited Environments},
  author={Ngo, Dat and Park, Hyun-Cheol and Kang, Bongsoon},
  journal={Electronics},
  volume={14},
  number={12},
  pages={2495},
  year={2025},
  publisher={MDPI}
}

@article{gholami2024ai,
  title={Ai and memory wall},
  author={Gholami, Amir and Yao, Zhewei and Kim, Sehoon and Hooper, Coleman and Mahoney, Michael W and Keutzer, Kurt},
  journal={IEEE Micro},
  volume={44},
  number={3},
  pages={33--39},
  year={2024},
  publisher={IEEE}
}

@article{shuvo2022efficient,
  title={Efficient acceleration of deep learning inference on resource-constrained edge devices: A review},
  author={Shuvo, Md Maruf Hossain and Islam, Syed Kamrul and Cheng, Jianlin and Morshed, Bashir I},
  journal={Proceedings of the IEEE},
  volume={111},
  number={1},
  pages={42--91},
  year={2022},
  publisher={IEEE}
}

@inproceedings{alexander2016focal,
  title={Focal flow: Measuring distance and velocity with defocus and differential motion},
  author={Alexander, Emma and Guo, Qi and Koppal, Sanjeev and Gortler, Steven and Zickler, Todd},
  booktitle={European conference on computer vision},
  pages={667--682},
  year={2016},
  organization={Springer}
}

@inproceedings{guo2017focal,
  title={Focal track: Depth and accommodation with oscillating lens deformation},
  author={Guo, Qi and Alexander, Emma and Zickler, Todd},
  booktitle={Proceedings of the IEEE international conference on computer vision},
  pages={966--974},
  year={2017}
}

@article{alexander2018focal,
  title={Focal flow: Velocity and depth from differential defocus through motion},
  author={Alexander, Emma and Guo, Qi and Koppal, Sanjeev and Gortler, Steven J and Zickler, Todd},
  journal={International Journal of Computer Vision},
  volume={126},
  number={10},
  pages={1062--1083},
  year={2018},
  publisher={Springer}
}

@article{guo2019compact,
  title={Compact single-shot metalens depth sensors inspired by eyes of jumping spiders},
  author={Guo, Qi and Shi, Zhujun and Huang, Yao-Wei and Alexander, Emma and Qiu, Cheng-Wei and Capasso, Federico and Zickler, Todd},
  journal={Proceedings of the National Academy of Sciences},
  volume={116},
  number={46},
  pages={22959--22965},
  year={2019},
  publisher={National Academy of Sciences}
}

@book{alexander2019theory,
  title={A theory of depth from differential defocus},
  author={Alexander, Emma},
  year={2019},
  publisher={Harvard University}
}

@inproceedings{alexander2021depth,
  title={Depth from defocus as a special case of the transport of intensity equation},
  author={Alexander, Emma and Kabuli, Leyla A and Cossairt, Oliver S and Waller, Laura},
  booktitle={2021 IEEE International Conference on Computational Photography (ICCP)},
  pages={1--13},
  year={2021},
  organization={IEEE}
}

@inproceedings{luo2025focal,
  title={Focal Split: Untethered Snapshot Depth from Differential Defocus},
  author={Luo, Junjie and Mamish, John and Fu, Alan and Concannon, Thomas and Hester, Josiah and Alexander, Emma and Guo, Qi},
  booktitle={Proceedings of the Computer Vision and Pattern Recognition Conference},
  pages={26965--26974},
  year={2025}
}

@article{luo2025depth,
  title={Depth from Coupled Optical Differentiation: J. Luo et al.},
  author={Luo, Junjie and Liu, Yuxuan and Alexander, Emma and Guo, Qi},
  journal={International Journal of Computer Vision},
  pages={1--18},
  year={2025},
  publisher={Springer}
}

@inproceedings{degalahal2005fpgapowermethodology,
  title={Methodology for high level estimation of FPGA power consumption},
  author={Degalahal, Vijay and Tuan, Tim},
  booktitle={Proceedings of the 2005 Asia and South Pacific Design Automation Conference},
  pages={657--660},
  year={2005}
}

@incollection{burt1987laplacian,
  title={The Laplacian pyramid as a compact image code},
  author={Burt, Peter J and Adelson, Edward H},
  booktitle={Readings in computer vision},
  pages={671--679},
  year={1987},
  publisher={Elsevier}
}

@article{anderson2004fpgapower,
  title={Power estimation techniques for FPGAs},
  author={Anderson, Jason Helge and Najm, Farid N},
  journal={IEEE Transactions on Very Large Scale Integration (VLSI) Systems},
  volume={12},
  number={10},
  pages={1015--1027},
  year={2004},
  publisher={Ieee}
}

@inproceedings{hu2018efficient,
  title={Efficient hardware architecture of softmax layer in deep neural network},
  author={Hu, Ruofei and Tian, Binren and Yin, Shouyi and Wei, Shaojun},
  booktitle={2018 IEEE 23rd International Conference on Digital Signal Processing (DSP)},
  pages={1--5},
  year={2018},
  organization={IEEE}
}

@inproceedings{geng2018hardware,
  title={Hardware-aware softmax approximation for deep neural networks},
  author={Geng, Xue and Lin, Jie and Zhao, Bin and Kong, Anmin and Aly, Mohamed M Sabry and Chandrasekhar, Vijay},
  booktitle={Asian Conference on Computer Vision},
  pages={107--122},
  year={2018},
  organization={Springer}
}

@article{ghose2018yourdram,
  title={What your DRAM power models are not telling you: Lessons from a detailed experimental study},
  author={Ghose, Saugata and Yaglik{\c{c}}i, Abdullah Giray and Gupta, Raghav and Lee, Donghyuk and Kudrolli, Kais and Liu, William X and Hassan, Hasan and Chang, Kevin K and Chatterjee, Niladrish and Agrawal, Aditya and others},
  journal={Proceedings of the ACM on Measurement and Analysis of Computing Systems},
  volume={2},
  number={3},
  pages={1--41},
  year={2018},
  publisher={ACM New York, NY, USA}
}

@manual{xilinxpowerestimator,
    title = {Xilinx Power Estimator User Guide},
    date = {2015-05-29},
    language = {English},
    version = {v2025.1},
    organization = {Xilinx},
    pubstate = {forthcoming},
}

@manual{latticepowerestimator,
    title = {Power Consumption and Management for ECP5 and ECP5-5G Devices},
    date = {2023-05-01},
    language = {English},
    version = {FPGA-TN-02210-1.4},
    organization = {Lattice},
    pubstate = {forthcoming},
}

@manual{hm0360manual,
    organization  = {Himax Technologies},
    title         = {Datasheet 1/6″ 640 x 480 VGA 60FPS CMOS Image Sensor},
    number        = {HM0360},
    year          =  {2020},
    month         =  {4}
}

@manual{ecp5evalmanual,
    organization = "Lattice Semiconductor",
    title        = "ECP5 Evaluation Board User Guide",
    number       = "FPGA-EB-02017-1.0",
    year         = 2018, 
    month = 7
}

@manual{ecp5familydatasheet,
    organization = {Lattice Semiconductor},
    title        = {ECP5 and ECP5-5G Family Data Sheet},
    number       = {FPGA-DS-02012-3.4},
    year         = {2025}, 
    month = {9}
}

@article{mamish2024nir,
  title={NIR-sighted: A programmable streaming architecture for low-energy human-centric vision applications},
  author={Mamish, John and Alharbi, Rawan and Sen, Sougata and Holla, Shashank and Kamath, Panchami and Sangar, Yaman and Alshurafa, Nabil and Hester, Josiah},
  journal={ACM Transactions on Embedded Computing Systems},
  volume={23},
  number={6},
  pages={1--26},
  year={2024},
  publisher={ACM New York, NY}
}

@inproceedings{kuon2006measuring,
  title={Measuring the gap between FPGAs and ASICs},
  author={Kuon, Ian and Rose, Jonathan},
  booktitle={Proceedings of the 2006 ACM/SIGDA 14th international symposium on Field programmable gate arrays},
  pages={21--30},
  year={2006}
}

@BOOK{hecht2017optics,
    author = {{Hecht}, Eugene},
    title = "{Optics}",
    year = 2017,
    adsurl = {https://ui.adsabs.harvard.edu/abs/2017opti.book.....H}
}

@inproceedings{zuo2025omni,
  title={Omni-dc: Highly robust depth completion with multiresolution depth integration},
  author={Zuo, Yiming and Yang, Willow and Ma, Zeyu and Deng, Jia},
  booktitle={Proceedings of the IEEE/CVF International Conference on Computer Vision},
  pages={9287--9297},
  year={2025}
}

@article{pentland1987new,
  title={A new sense for depth of field},
  author={Pentland, Alex Paul},
  journal={IEEE transactions on pattern analysis and machine intelligence},
  number={4},
  pages={523--531},
  year={1987},
  publisher={IEEE}
}

@article{levin2007image,
  title={Image and depth from a conventional camera with a coded aperture},
  author={Levin, Anat and Fergus, Rob and Durand, Fr{\'e}do and Freeman, William T},
  journal={ACM transactions on graphics (TOG)},
  volume={26},
  number={3},
  pages={70--es},
  year={2007},
  publisher={ACM New York, NY, USA}
}

@article{watanabe1998rational,
  title={Rational filters for passive depth from defocus},
  author={Watanabe, Masahiro and Nayar, Shree K},
  journal={International Journal of Computer Vision},
  volume={27},
  number={3},
  pages={203--225},
  year={1998},
  publisher={Springer}
}

@article{zhou2011coded,
  title={Coded aperture pairs for depth from defocus and defocus deblurring},
  author={Zhou, Changyin and Lin, Stephen and Nayar, Shree K},
  journal={International journal of computer vision},
  volume={93},
  number={1},
  pages={53--72},
  year={2011},
  publisher={Springer}
}

@inproceedings{mirdehghan2018optimal,
  title={Optimal structured light a la carte},
  author={Mirdehghan, Parsa and Chen, Wenzheng and Kutulakos, Kiriakos N},
  booktitle={Proceedings of the IEEE conference on computer vision and pattern recognition},
  pages={6248--6257},
  year={2018}
}

@inproceedings{o20143d,
  title={3D shape and indirect appearance by structured light transport},
  author={O'Toole, Matthew and Mather, John and Kutulakos, Kiriakos N},
  booktitle={Proceedings of the IEEE Conference on Computer Vision and Pattern Recognition},
  pages={3246--3253},
  year={2014}
}

@article{puglia2017real,
  title={Real-time low-power FPGA architecture for stereo vision},
  author={Puglia, Luca and Vigliar, Mario and Raiconi, Giancarlo},
  journal={IEEE Transactions on Circuits and Systems II: Express Briefs},
  volume={64},
  number={11},
  pages={1307--1311},
  year={2017},
  publisher={IEEE}
}

@article{lu2021resource,
  title={A resource-efficient pipelined architecture for real-time semi-global stereo matching},
  author={Lu, Zhimin and Wang, Jue and Li, Zhiwei and Chen, Song and Wu, Feng},
  journal={IEEE Transactions on Circuits and Systems for Video Technology},
  volume={32},
  number={2},
  pages={660--673},
  year={2021},
  publisher={IEEE}
}

@article{ttofis2015low,
  title={A low-cost real-time embedded stereo vision system for accurate disparity estimation based on guided image filtering},
  author={Ttofis, Christos and Kyrkou, Christos and Theocharides, Theocharis},
  journal={IEEE Transactions on Computers},
  volume={65},
  number={9},
  pages={2678--2693},
  year={2015},
  publisher={IEEE}
}

@article{zhang2018nipm,
  title={NIPM-sWMF: Toward efficient FPGA design for high-definition large-disparity stereo matching},
  author={Zhang, Xuchong and Sun, Hongbin and Chen, Shiqiang and Song, Lin and Zheng, Nanning},
  journal={IEEE Transactions on Circuits and Systems for Video Technology},
  volume={29},
  number={5},
  pages={1530--1543},
  year={2018},
  publisher={IEEE}
}

@article{jin2014fast,
  title={Fast and accurate stereo vision system on FPGA},
  author={Jin, Minxi and Maruyama, Tsutomu},
  journal={ACM Transactions on Reconfigurable Technology and Systems (TRETS)},
  volume={7},
  number={1},
  pages={1--24},
  year={2014},
  publisher={ACM New York, NY, USA}
}

@inproceedings{mattoccia2015passive,
  title={A passive RGBD sensor for accurate and real-time depth sensing self-contained into an FPGA},
  author={Mattoccia, Stefano and Poggi, Matteo},
  booktitle={Proceedings of the 9th International Conference on Distributed Smart Cameras},
  pages={146--151},
  year={2015}
}

@article{joseph2014video,
  title={Video-rate calculation of depth from defocus on a FPGA},
  author={Joseph Raj, Alex Noel and Staunton, Richard C},
  journal={Journal of Real-Time Image Processing},
  volume={14},
  number={2},
  pages={469--480},
  year={2014},
  publisher={Springer}
}

@article{wang2022block,
  title={A block patchmatch-based energy-resource efficient stereo matching processor on FPGA},
  author={Wang, Hongyu and Zhou, Wei and Zhang, Xiangyu and Lou, Xin},
  journal={IEEE Transactions on Circuits and Systems I: Regular Papers},
  volume={69},
  number={7},
  pages={2893--2905},
  year={2022},
  publisher={IEEE}
}
}

\newcounter{si}
\setcounter{si}{0} 
\renewcommand\thesection{S\arabic{si}}
\newcounter{fi}
\setcounter{fi}{0} 
\renewcommand{\thefigure}{S\arabic{fi}}
\newcounter{ti}
\setcounter{ti}{0} 
\renewcommand{\thetable}{S\arabic{ti}}

\clearpage
\setcounter{page}{1}
\maketitlesupplementary

\stepcounter{si}
\section{Implementation Details}

In this section, we provide implementation details of the five steps to the SpiderCam algorithm.

\stepcounter{fi}
\begin{figure}[h!]
    \centering
    \includegraphics[width=\linewidth]{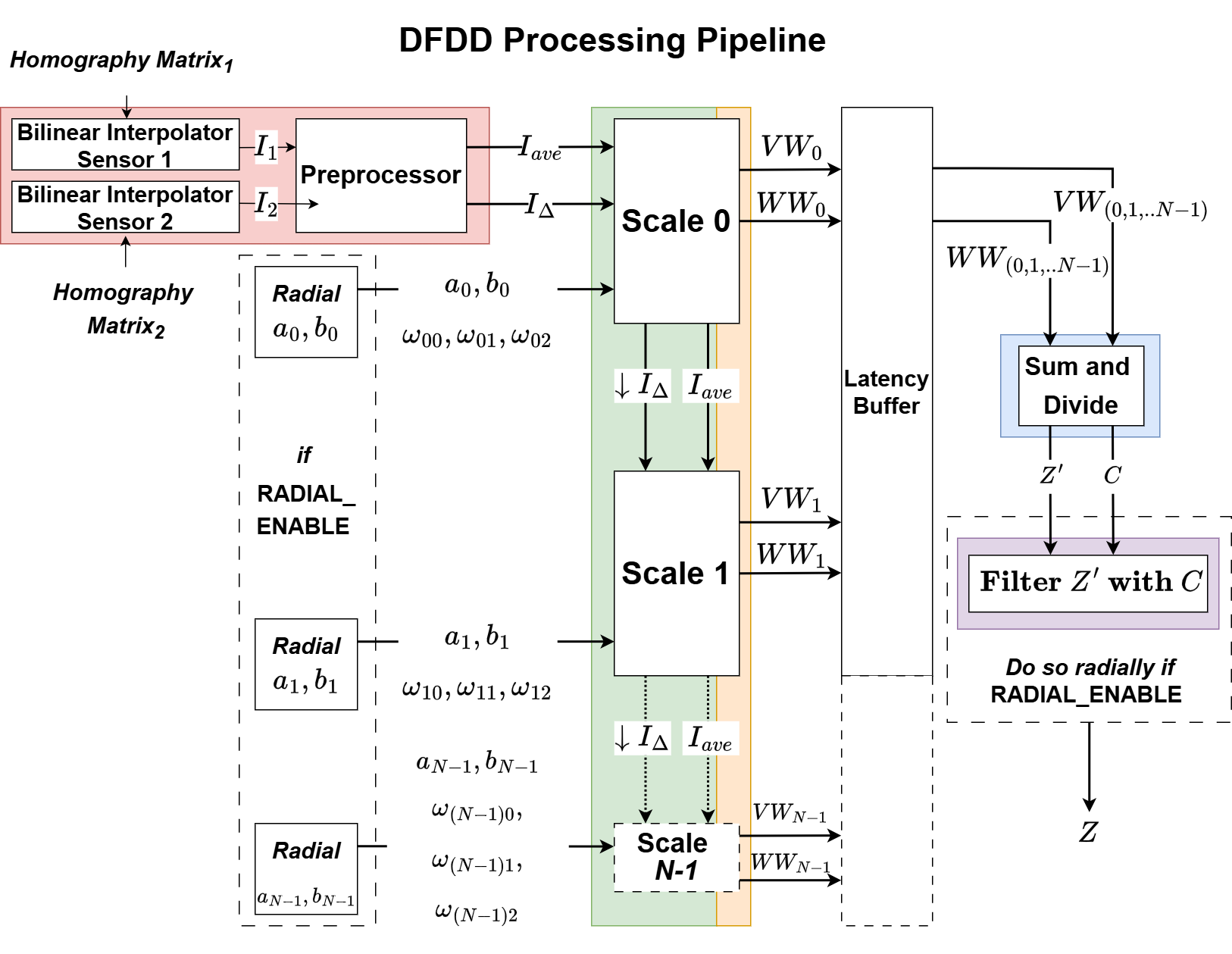}
    \caption{General overview of the algorithm.}
\end{figure}

\subsection{Homography}
We apply an affine transformation homography using bilinear interpolation to the images to ensure sub-pixel alignment.  
A purely mechanical solution to align the images would be expensive and difficult to do due to the high degree of precision required.  
The closer focal plane image is known as \( I_{1} \), and the further focal plane image as \( I_{2} \).

\subsection{Preprocessing}

\stepcounter{fi}
\begin{figure}[h!]
    \centering
    \includegraphics[width=\linewidth]{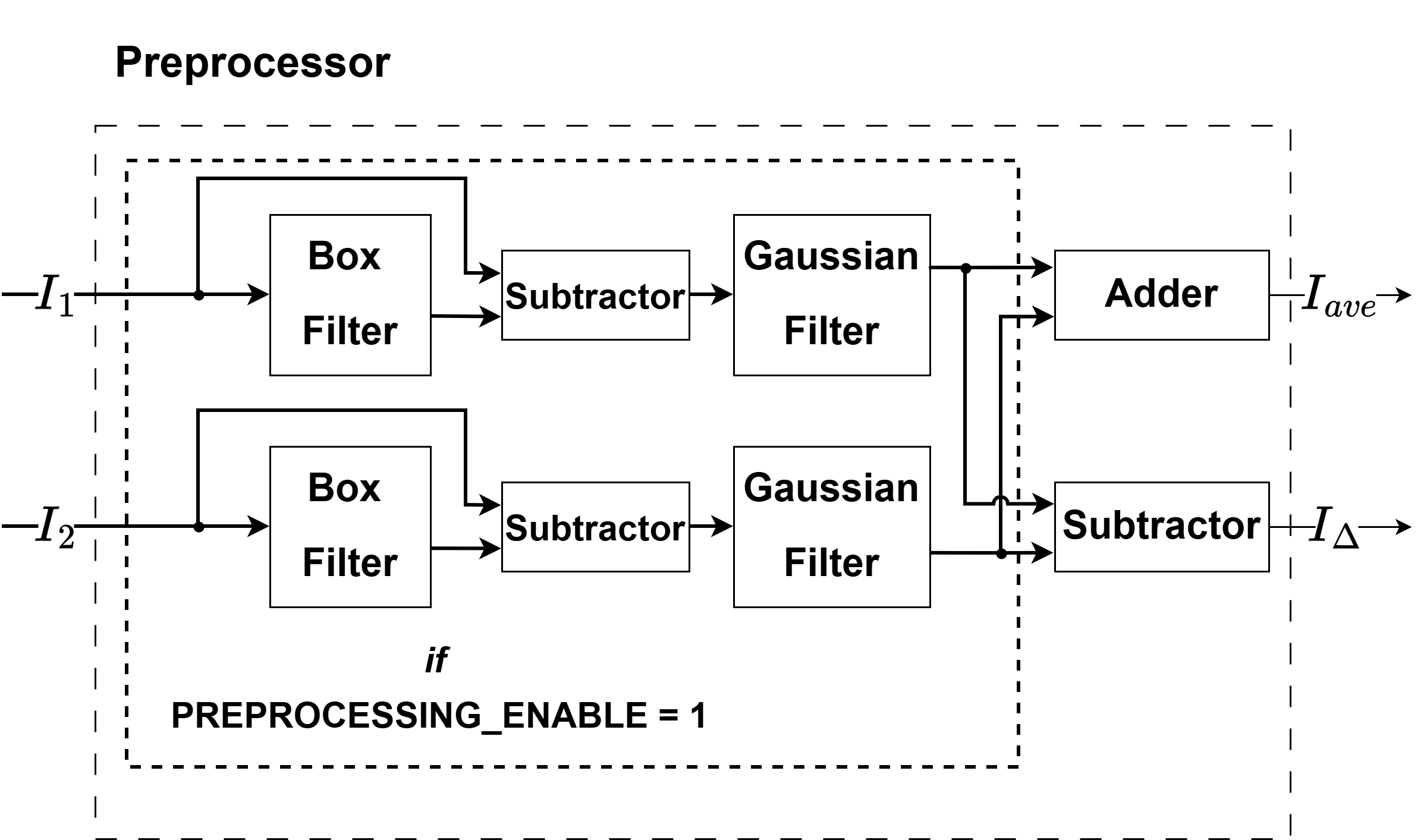}
    \caption{Preprocessing module.}
\end{figure}

In the preprocessing step, to denoise the image, we apply a box filter followed by subtraction and then a Gaussian filter.  
This step only applies if \texttt{PREPROCESSING\_ENABLE} is set to one.  
Otherwise, we skip directly to calculating \( I_{ave} \) and \( I_{\Delta} \), which are simply the sum and difference of \( I_{1} \) and \( I_{2} \):

\[
I_{ave} = I_1 + I_2, \qquad
I_{\Delta} = I_1 - I_2
\]

\subsection{$N^{th}$ Scale}

\stepcounter{fi}
\begin{sidewaysfigure}
    \centering
    \includegraphics[width=\textheight]{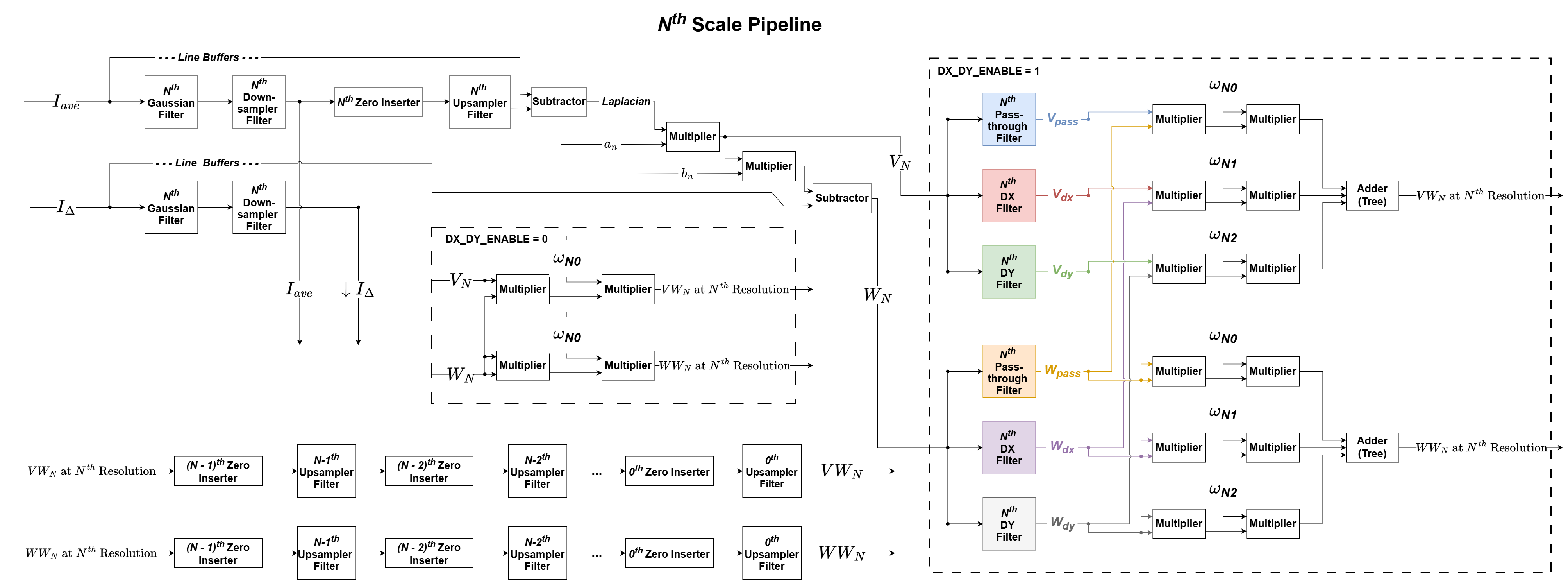}
    \caption{Computation for the $N^{th}$ scale.}
    \label{fig:$N^{th}$_scale}
\end{sidewaysfigure}

For a particular scale \( N \), we define:
\[
V_{N} = a_n \times\texttt{laplacian}(I_{ave})   
\]
\[
W_{N} =  b_n \times V_{N}  - I_{\Delta}
\]

\textbf{Then, without additional spatial derivatives:}
\[
VW_N = w_{n0} \times V_{N} \times W_{N}
\]
\[
WW_N = w_{n0} \times W_{N} \times W_{N}
\]

\textbf{With additional spatial derivatives enabled} (as indicated by \texttt{DX\_DY\_ENABLE}), 
we sum the derivatives to produce \( VW_N \) and \( WW_n \):

\[
\begin{aligned}
VW_N = &\; w_{n0} \times  V_{N} \times W_{N} \\
      &+  w_{n1} \times V_{N\text{ }}dx \cdot W_{N\text{ }}dx  \\
      &+ w_{n2} \times V_{N\text{ }}dy \times W_{N\text{ }}dy 
\end{aligned}
\]

\[
\begin{aligned}
WW_N = &\; w_{n0} \times  W_{N} \times W_{N} \\
      &+  w_{n1} \times W_{N\text{ }}dx \cdot W_{N\text{ }}dx  \\
      &+ w_{n2} \times W_{N\text{ }}dy \times W_{N\text{ }}dy 
\end{aligned}
\]

\subsection{Summation and Division}
The final step is to sum the respective $VW$s and $WW$s from each scale and divide, to give the depth and confidence values:
\[
Z' = \frac{\displaystyle \sum_{i=0}^{N-1} VW_N}{\displaystyle \sum_{i=0}^{N-1} WW_N},
\qquad
C = \sum_{i=0}^{N-1} WW_N
\]

The resulting \( Z' \) in then filtered using the confidence value and working range values to produce the final depth map \( Z \).

\subsection{Radial Zone Adaptation}
\label{subsec:rz_adapt_supp}
When \texttt{RADIAL\_ENABLE} is active, the logic for per-zone parameter selection operates as follows:

\begin{algorithm}[H]
    \caption{\textbf{Spatial variation.} Zone-based parameter evaluation uses squared distance to avoid hardware square-root operations and tracks pixel streaming to save memory.}
    \label{alg:radial_zone_selection_supp}
    \begin{algorithmic}[1]
        \STATE Input: pixel's distance from center, \\ $streaming\_distance\_square$, \\ per zone and/or per scale values, \\ $ r^2_{zones}[~~],a'_{scales}[~~], b'_{scales}[~~], C'_{thresh}[~~], Z'_{min}[~~], Z'_{max}[~~]$ \\
        \STATE Output: parameters $(\{a_0,...,a_{N-1}\},\{b_0,...,b_{N-1}\},$ \\ 
        $\{\omega_0, ..., \omega_{(3N - 1)}\}$, $C_{thresh}, Z_{min}, Z_{max})$ \\
        \FOR{$z$ in $\{zones - 1, zones - 2, ... , 0\}$}
            \IF{$streaming\_distance\_squared < r^2_{zone}[z]$}
                \FOR{$scale$ in \{0,1,..,$N-1$\}}
                    \STATE $a_{scale} \gets a'_{scales}[z]$
                    \STATE $b_{scale} \gets b'_{scales}[z]$
                \ENDFOR
                \STATE $C_{thresh} \gets C'_{thresh}[z]$
                \STATE $Z_{min} \gets Z'_{min}[z]$
                \STATE $Z_{max} \gets Z'_{max}[z]$
            \ENDIF
        \ENDFOR
    \end{algorithmic}
\end{algorithm}
A single confidence minimum value can be used to filter the depth map but due to the Petzval field curvature, exacerbated by using a small focal length and a very small image sensor, our working range becomes quite limited. To handle this, we assign a set of parameters not only for each scale, but also for each \textbf{radial zone} that the pixel belongs to. Using the pixel's \((x, y)\) coordinate, the $streaming\_distance\_squared$ can be computed, which is simply the distance squared of the pixel from the optical center of the image.

\stepcounter{si}
\section{Image Stream Processing in Hardware}

In hardware, to process images efficiently, we adopt a \emph{data streaming model}. 
This is a standard approach for maximizing \emph{data locality} and works well with our algorithm since the computations apply only to local pixels. 
Instead of buffering an entire image frame, each convolution filter requires the kernel's height minus one lines to be buffered.

\subsection{Use of Efficient Gaussian Kernel}
\label{sec:efficient_kernels}
Our Gaussian kernel is the $5 \times 5$ Anderson--Burt Gaussian kernel \cite{burt1987laplacian}:

\[
\frac{1}{256}
\begin{bmatrix}
1 & 4 & 6 & 4 & 1 \\
4 & 16 & 24 & 16 & 4 \\
6 & 24 & 36 & 24 & 6 \\
4 & 16 & 24 & 16 & 4 \\
1 & 4 & 6 & 4 & 1
\end{bmatrix}
\]

which can also be expressed in its linearly separable form as:

\[
\frac{1}{16}
\begin{bmatrix}
1 & 4 & 6 & 4 & 1
\end{bmatrix}
\otimes
\frac{1}{16}
\begin{bmatrix}
1 \\[3pt] 4 \\[3pt] 6 \\[3pt] 4 \\[3pt] 1
\end{bmatrix}
\]

This kernel is both compact—using small integer coefficients that are powers of two—and linearly separable, 
making it highly efficient for implementation in hardware.

\stepcounter{si}
\section{Hybrid Fixed Point to Efficient Floating Point Model}
\label{sec:hf_fp_supp}
To minimize resource usage, the homography and preprocessing stages are implemented using \emph{fixed-point arithmetic}. 
In particular, both the box filter and Gaussian filter are executed without trimming bits, thereby avoiding quantization issues. 
After these stages, the results are converted to floating point before computing $I_{ave}$ and $I_{\Delta}$.

On the floating-point side, two main strategies make the design feasible:

\begin{itemize}
    \item Use of \textbf{half precision (FP16)} instead of full precision (FP32)
    \item \textbf{Dropping subnormal number support}
\end{itemize}

Performing floating-point arithmetic with subnormal number support makes multipliers and dividers nearly as expensive as fixed-point arithmetic, with only a small additional cost for exponent handling. 
With subnormal support disabled, we can assume that the most significant bit (MSB) of both operands is always one. 
This guarantees that the MSB of the result will appear in either the highest or second-highest bit position, eliminating the need for an expensive variable shifter during multiplication or division.

Our adders still require the variable shifters and priority encoders associated with floating-point arithmetic, 
but some of the surrounding logic is simplified by removing subnormal handling. 
The primary reduction in resource usage for our floating-point adders comes from using half-precision (FP16) representation.

\stepcounter{si}
\section{The General Upsampling Problem}
\label{sec:upsampling_supp}
Our algorithm requires the generation of a \emph{Laplacian Pyramid} in the $N^{th}$ Scale Pipeline. 
Downsampling an image is straightforward; however, a naïve implementation of upsampling using bilinear interpolation would require buffering the downsampled image, which corresponds to one-quarter the number of lines of the original image. 
This occurs because pixels that are reverse-mapped from the upsampled image are far apart but refer to the same neighboring pixels in the downsampled image. 
This can be described as:
\[
f(x, y) = g(x // 2,\, x // 2 + 1,\, y // 2,\, y // 2 + 1)
\]

A more efficient solution leverages the fact that the downsampled image dimensions are halved. 
We can therefore push out a pixel from the downsampled image every four clock cycles for each clock cycle in the upsampled image. 
This reduces the number of lines that must be buffered to one-half of the downsampled image, which corresponds to one-eighth the number of lines of the original image. 
Some buffering is still required, since every other line in the upsampled image uses the same neighboring pixels from the downsampled image. 
Although buffering one-eighth is an improvement, it remains unfeasible (for example, one-eighth of a 480-line image is 60 lines) and scales undesirably with the image height.

\subsection{Proposed Solution}
First, a clarification on how the convolutions are performed.
For odd-sized kernels, which are the majority of our kernels, we crop
evenly from all sides of the image to retain the original image dimension.

For even-sized kernels, due to ambiguity of cropping, we specify which border
we crop from to get back to the original image dimension. In hardware, we don't
actually do cropping but instead adjust the 'center' of the kernel to achieve
the same behavior. We define 'br' as the bottom and right crop, and 'tl' as the top and left crop.

We solve the problem generally by avoiding the need to buffer half an image. Instead, we double the kernel sizes and interleave their coefficients with zeros in conjunction with inserting zeros into the image itself. 
This can be summarized using the following formulations.

\paragraph{Naïve approach:}
\begin{enumerate}
    \item Image is convolved with a $2 \times 2$ box filter, known as our 'Downsampler Filter'.
    \item Skip every odd row and column.
    \item The resulting downsampled image is then upsampled using buffering and a bilinear interpolation.
\end{enumerate}

\noindent
\textbf{Buffering requirement:} one-quarter of the image lines.

\paragraph{Our approach:}
\begin{enumerate}
    \item Image is convolved with a $2 \times 2$ box filter, known as our 'Downsampler Filter'
    \item Insert zero at the pixels with odd row and column coordinates (zero insertion).
    \item The zero-inserted image is convolved with a bilinear interpolation kernel, known as our 'Upsampler Filter'.
\end{enumerate}

With the bilinear interpolation kernel specified, the upsampling process can be expressed as:

\begin{align*}
    \text{Image} * 
    \begin{bmatrix}
    \frac{1}{4} & \frac{1}{4} \\
    \frac{1}{4} & \frac{1}{4}
    \end{bmatrix}_{\text{tl}}
    &\;\rightarrow\;
    \text{zero insert} \\[4pt]
    &\;\rightarrow\;
    *
    \begin{bmatrix}
    \frac{1}{4} & \frac{1}{2} & \frac{1}{4} \\
    \frac{1}{2} & 1 & \frac{1}{2} \\
    \frac{1}{4} & \frac{1}{2} & \frac{1}{4}
    \end{bmatrix}
    \;\rightarrow\;
    \text{upsampled image}
\end{align*}

Because the box filter is even-sized, it introduces a \emph{half-pixel shift} with respect to the coordinate space of $I_{\Delta}$. 
We correct this shift by convolving further with a half-pixel-shifted kernel using the \emph{opposite cropping side}:

\begin{align*}
\text{Image} * 
\begin{bmatrix}
\frac{1}{4} & \frac{1}{4} \\
\frac{1}{4} & \frac{1}{4}
\end{bmatrix}_{\text{tl}}
&\;\rightarrow\;
\text{zero insert} \\[4pt]
&\;\rightarrow\;
*
\begin{bmatrix}
\frac{1}{4} & \frac{1}{2} & \frac{1}{4} \\
\frac{1}{2} & 1 & \frac{1}{2} \\
\frac{1}{4} & \frac{1}{2} & \frac{1}{4}
\end{bmatrix} \\[4pt]
&\;\rightarrow\;
*
\begin{bmatrix}
\frac{1}{4} & \frac{1}{4} \\
\frac{1}{4} & \frac{1}{4}
\end{bmatrix}_{\text{br}}
\;\rightarrow\;
\text{upsampled image}
\end{align*}

The convolution of the two kernels,

\[
\begin{bmatrix}
\frac{1}{4} & \frac{1}{2} & \frac{1}{4} \\
\frac{1}{2} & 1 & \frac{1}{2} \\
\frac{1}{4} & \frac{1}{2} & \frac{1}{4}
\end{bmatrix}
*
\begin{bmatrix}
\frac{1}{4} & \frac{1}{4} \\
\frac{1}{4} & \frac{1}{4}
\end{bmatrix},
\]

produces:

\[
\begin{bmatrix}
1 & 3 & 3 & 1 \\
3 & 9 & 9 & 3 \\
3 & 9 & 9 & 3 \\
1 & 3 & 3 & 1
\end{bmatrix}_{\text{br}} \times \frac{1}{16}
\]

which is \emph{linearly separable} into:

\[
\begin{bmatrix} 1 & 3 & 3 & 1 \end{bmatrix}_{\text{br}} 
\times \frac{1}{4} 
\times
\begin{bmatrix}
1 \\[2pt] 3 \\[2pt] 3 \\[2pt] 1
\end{bmatrix}_{\text{br}} 
\times \frac{1}{4}.
\]

\noindent
\textbf{Buffering requirement:} three lines of the image.

Now with this solution, at the $0^{th}$ scale, it is simple: we downsample, then upsample, and proceed as normal. 
But for the $1^{st}$ scale, where $\downarrow_1I_{ave}$ and $\downarrow_1I_{\Delta}$ are used, we must interleave the kernels with zeros to match the appropriate dimensions.

For example, the $N^{th}$ Anderson--Burt Gaussian kernel across scales is given below.

\vspace{6pt}

\noindent{$0^{th}$ scale}\\[3pt]
\[
\frac{1}{16}
\begin{bmatrix}
1 & 4 & 6 & 4 & 1
\end{bmatrix}
\]

\vspace{6pt}

\noindent{$1^{st}$ scale}\\[3pt]
\[
\frac{1}{16}
\begin{bmatrix}
1 & 0 & 4 & 0 & 6 & 0 & 4 & 0 & 1
\end{bmatrix}
\]

\vspace{6pt}

\noindent{$2{nd}$ scale}\\[3pt]
\resizebox{\linewidth}{!}{$\frac{1}{16}
\left[\begin{array}{@{}*{17}{r}@{}}
1 & 0 & 0 & 0 & 4 & 0 & 0 & 0 & 6 & 0 & 0 & 0 & 4 & 0 & 0 & 0 & 1
\end{array}\right]$}

\vspace{10pt}

The $N^{th}$ zero-inserter follows:
\[
f(x,y) =
\begin{cases}
\parbox[t]{0.75\linewidth}{
\centering keep original pixel, if $x \bmod 2^{(N^{th}+1)} = 0$ and $y \bmod 2^{(N^{th}+1)} = 0$
},\\[8pt]
\parbox[t]{0.75\linewidth}{
\centering 0, otherwise.
}
\end{cases}
\]

The power-of-two modulus makes this operation straightforward to implement in hardware. 

To then reconstruct the full-resolution image, chain zero-inserters and upsamplers from each scale back to the original size, as shown in \cref{fig:$N^{th}$_scale}. 

\stepcounter{si}
\section{FLOPs per Pixel}

All the $N$th filters are equivalent besides the interleaved zeros, which make them sparse kernels to match their scale dimension. 
This means that the adder tree sizes and number of multipliers between scales are the same, since all the zero values can simply be optimized away. 

As discussed previously, the majority of the kernels are linearly separable and consist of coefficients that are powers of two. 
Therefore, in the following calculations:
\begin{itemize}
    \item A multiply by a power of two is denoted as an \textbf{easy multiply}.
    \item A multiply requiring a full multiplier is denoted as a \textbf{true multiply}.
\end{itemize}

Here, although $N$ refers to the total number of scales, it can also refer to a particular scale if noted within the context of a scale and ``$N^{th}$'' always refers to a particular scale (starting from $0$).  
Please refer to \cref{fig:$N^{th}$_scale} for visual reference and tables \cref{tab:flops_per_op} to see FLOPs per Filter/Operation and \cref{tab:flops_agg} Aggregate costs for N Scales.

\stepcounter{ti}
\begin{table*}[t]
\centering
\renewcommand{\arraystretch}{1.2}
\setlength{\tabcolsep}{4pt}
\begin{tabular}{|l|c|c|c|c|c|}
\hline
\textbf{Filter / Operation} & \textbf{Adders} & \textbf{True Mult.} & \textbf{Easy Mult.} & \textbf{Dividers} & \textbf{Total FLOPs} \\ \hline
Gaussian $5\times5$ Filter & 8 & 2 & 8 & 0 & 18 \\ \hline
Downsampler Filter & 2 & 0 & 4 & 0 & 6 \\ \hline
Upsampler Filter & 6 & 4 & 4 & 0 & 14 \\ \hline
$V_{N}$ \& $W_{N}$ & 2 & 2 & 0 & 0 & 4 \\ \hline
Cross Mult. w/ Weights (DX\_DY\_ENABLE = 1) & 8 & 12 & 10 & 0 & 30 \\ \hline
Cross Mult. w/ Weights (DX\_DY\_ENABLE = 0) & 0 & 4 & 0 & 0 & 4 \\ \hline
Add and Divide & $2 \times (N - 1)$ & 0 & 0 & 1 & $2 \times (N - 1) + 1$ \\ \hline
\end{tabular}
\caption{FLOP counts and arithmetic breakdown per operation. $N$ refers to the number of scales. Total FLOPs is the sum of all arithmetic operations per filter or stage.}
\label{tab:flops_per_op}
\end{table*}

\vspace{8pt}
\noindent\textbf{Computation Breakdown per Scale}

An $N^{\text{th}}$ scale requires:
\begin{itemize}
    \item $2 \times$ Gaussian Filter
    \item $2 \times$ Downsampler Filter
    \item $1 + 2 \times $ Upsampler Filter
    \item $V_N$ \& $W_N$
    \item $1 \times$ Cross Multiplication with Weights (depending on DX\_DY\_ENABLE)
\end{itemize}

\stepcounter{ti}
\begin{table*}[t]
\centering
\renewcommand{\arraystretch}{1.2}
\setlength{\tabcolsep}{4pt}
\begin{tabular}{|c|c|c|c|c|c|c|}
\hline
\textbf{N (Scales)} & \textbf{DX\_DY\_ENABLE} & \textbf{Adders} & \textbf{True Mult.} & \textbf{Easy Mult.} & \textbf{Dividers} & \textbf{Total FLOPs} \\ \hline
1 & 0 & 28  & 14 & 28  & 1 & \textbf{71}  \\ 
1 & 1 & 36  & 22 & 38  & 1 & \textbf{97}  \\ \hline
2 & 0 & 70  & 36 & 64  & 1 & \textbf{171} \\ 
2 & 1 & 86  & 52 & 84  & 1 & \textbf{223} \\ \hline
3 & 0 & 124 & 66 & 108 & 1 & \textbf{299} \\ 
3 & 1 & 148 & 90 & 138 & 1 & \textbf{377} \\ \hline
\end{tabular}
\caption{Aggregate per-pixel operation counts across $N$ scales using the per-operation table and the updated per-scale rule 
(2× Gaussian, 2× Downsampler, $(1+2\times N^{th})$ Upsamplers per scale, and 1× Cross Mult. with Weights per scale),$V_N \text{ \& } W_N$ plus one final Add and Divide.}
\label{tab:flops_agg}
\end{table*}

\stepcounter{si}
\section{Memory Usage}

The number of image lines that must be buffered depends on the kernel height and the amount of zero interleaving at each scale.  
Here are the line-buffer equations for each buffer type at an $N^{th}$ scale (where $N^{th}$ = 0, 1, 2, \dots N-1).  
$G_h$, $D_h$, $U_h$, and $P_h$ denote the base kernel heights of the Gaussian, Downsampler, Upsampler, and Pass/DX/DY filters, respectively.

\vspace{6pt}
\noindent\textbf{$N^{th}$ Scale Buffer Equations}
\[
\begin{aligned}
\text{$N^{th}$ Gaussian } (5\times5): &\quad (G_h - 1)\,2^{N^{th}} = 2^{N^{th} + 2} \\[3pt]
\text{$N^{th}$ Downsampler } (2\times2): &\quad 2^{N^{th}} \\[3pt]
\text{$N^{th}$ Upsampler } (4\times4): &\quad 3 \times 2^{N^{th}} \\[3pt]
\text{$N^{th}$ Pass/DX/DY } (3\times3): &\quad 2^{N^{th} + 1}
\end{aligned}
\]

\vspace{6pt}
\noindent\textbf{$N^{th}$ Scale Buffers, $B(N^{th})$}

An $N^{\text{th}}$ scale requires these many buffers:
\begin{itemize}
    \item $4 \times$ $N^{th}$ Gaussian buffers  
    \item $4 \times$ $N^{th}$ Downsampler buffers  
    \item $3 \times$ $N^{th}$ Upsampler buffers  
    \item $2 \times$ $N^{th}$ Pass/DX/DY buffers (only if DX\_DY\_ENABLE = 1)  
    \item $2 \times \displaystyle\sum_{i=0}^{N-1} \big[\min(i, 1) \times (i - 1)^{th}\ \text{Upsampler buffer}\big]$
\end{itemize}
Which simplifies to:
\begin{itemize}
    \item $33 \times 2^{N^{th}} + \\ 
    2 \times \sum_{i=0}^{N-1}
    [\min(i, 1) \times (i - 1)^{th}\ \text{Upsampler buffer}] \\
    \text{(DX\_DY\_ENABLE = 1)}$ 
    \\
    \item $29 \times 2^{N^{th}} + \\
    2 \times \sum_{i=0}^{N-1}
    [\min(i, 1) \times (i - 1)^{th}\ \text{Upsampler buffer}] \\
    \text{(DX\_DY\_ENABLE = 0)}$
\end{itemize}

Since each scale has a different number of buffers, and higher scales begin processing only after receiving their downsampled $I_{ave}$ and $I_{\Delta}$,  
they reach the latency buffer at different times. Thus, the latency buffer must store data equal to the longest latency path (largest scale) minus the shortest latency path (lowest scale), multiplied by $(N - 1)$ scales.

\vspace{6pt}
\noindent\textbf{$N^{th}$ Scale Latency, $L(N^{th})$}

An $N^{\text{th}}$ scale latency amounts to these number of buffers:
\begin{itemize}
    \item $1 \times$ Gaussian buffer  
    \item $1 \times$ Downsampler buffer  
    \item $1 \times$ Upsampler buffer  
    \item $1 \times$ Pass/DX/DY buffer (if DX\_DY\_ENABLE = 1)
    \item $\displaystyle\sum_{i=0}^{N-1} \big[\min(i, 1) \times (i - 1)^{th}\ \\ \text{Upsampler \& Downsampler Buffer}\big]$
\end{itemize}
\noindent
Which simplifies to,
\begin{itemize}
    \item $10 \times 2^{N^{th}} + \\
    \sum_{i=0}^{N-1}
    [\min(i, 1) \times (i - 1)^{th}\ \\ \text{Upsampler \& Downsampler buffer}] \\
    \text{(DX\_DY\_ENABLE = 1)}$ 
    \\
    \item $8 \times 2^{N^{th}} + \\
    \sum_{i=0}^{N-1}
    [\min(i, 1) \times (i - 1)^{th}\ \\ \text{Upsampler \& Downsampler buffer}] \\
    \text{(DX\_DY\_ENABLE = 0)}$ 
\end{itemize}

\noindent The latency buffer therefore needs to buffer:
\[
(N - 1) \times \big(L(N-1) - L(0)\big)
\]
lines.

See \cref{tab:lines_buffered} for the aggregate number of buffered lines across scales.

\stepcounter{ti}
\begin{table*}[t]
\centering
\renewcommand{\arraystretch}{1.3}
\setlength{\tabcolsep}{4pt}
\begin{tabular}{|c|c|c|c|c|}
\hline
\textbf{N (Scales)} & \textbf{DX\_DY\_ENABLE} & \textbf{Scale Buffers} & \textbf{Latency Buffers} & \textbf{Total Buffers} \\ \hline
1 & 0 & 29  & 0  & 29  \\ 
1 & 1 & 33  & 0  & 33 \\ \hline
2 & 0 & 93  & 12  &  105\\ 
2 & 1 & 105 & 14 & 119 \\ \hline
3 & 0 & 227 & 36  & 263\\ 
3 & 1 & 255 & 42 & 297 \\ \hline
\end{tabular}
\caption{Aggregate buffered line requirements for $N = 1$--$3$ scales, including buffers used in scales, buffers consumed by Latency Buffer, and Total buffers used. Each buffer is the size of one line of the image in FP16. For both DX\_DY\_ENABLE modes.}
\label{tab:lines_buffered}
\end{table*}

\stepcounter{si}
\section{Sensor Selection}
\label{sec:suppsensor}
To maintain low power usage, we selected the Himax HM0360 Ultra Low Power sensor. It is quite small, measuring at 2.88 mm across its diagonal and with a pixel pitch of 3.6 $\mu$m. We calculated that running at 32.5 FPS, the sensor only draws between 10-20 mW. The tradeoffs are the increased amount of noise in the image and using a smaller focal length to maintain a similar FoV. Using a shorter focal length creates a stronger field curvature, which we fix with our spatially varying parameters discussed earlier.

\stepcounter{si}
\section{FPGA power consumption and resource utilization estimation}
\label{sec:power_estimates_supp}
Estimating power consumption for a previous work without access to its source code is difficult: the only way to achieve a precise estimate of the power consumed by an FPGA design is to simulate the design on real data and record net switching information, using these figures for calculation~\cite{anderson2004fpgapower}. Although this is not feasible for many existing works as designs and data are not readily available, we can use vendor tools~\cite{xilinxpowerestimator,latticepowerestimator} to attain a reasonably accurate power estimate if we are provided with circuit operating frequency, resource utilization, fanout, and activity factor~\cite{degalahal2005fpgapowermethodology}. While frequency and resource utilization are readily available from the papers themselves, fanout (the average number of loads driven by a driver, e.g. the average number of LUTs driven by a flip-flop) cannot be known without the actual design and activity factor (number of wire toggles between '0' and '1' per clock cycle) requires the design to be simulated on actual data. 

Therefore, to get a power estimate, we choose low and high reasonable values for fanout and activity factor. For fanout, we simply choose values of 3 and 6 for low- and high- power estimates, respectively~\cite{xilinxpowerestimator}.

"Activity factor" (AF) refers to the number of times a logic signal toggles from '0' to '1' and back to '0' during a single clock cycle. A system's power consumption is very strongly affected by its AF; power dissipation in digital circuits largely comes from charging and discharging internal nodes. AF strongly depends on input data: if input data doesn't change from cycle to cycle, then there is no change in internal logic signals and the AF is 0\%. Neglecting glitching, the worst-case AF is 50\%. For random signals, the AF will be 25\%. ~\cite{xilinxpowerestimator} recommends using an AF of 12.5\% for conservatively estimating power on real-world data. In our table, we use an AF of 12.5\% for our low-end power estimate and an AF of 50\% for our high-end power estimate. Furthermore, we normalize all circuits to operate at 30 Mpix/s, roughly the number of pixels per second needed to process an HD video frame at 30 FPS, scaling the AF by the fraction of clock cycles that the system has to be active for to achieve 30 Mpix/s. For instance, if a work is capable of processing 100 Mpix/s, we scale the activity factor by 30 / 100, modeling reduced power consumption during inter-frame idle time due to an AF of 0\%.

A more detailed version of~\cref{tab:powertable} can be seen in~\cref{tab:detailed_powertable}. The ``Estimated Core Power" column shows power estimates reported in the original papers, where available. In the original works, these estimates were produced not by measuring the power consumption of a running system, but by feeding their design into a vendor-provided tool as described above. Because of differences in frame-rate and estimation methodology, the reported power numbers cannot be equitably compared, so we re-did the estimates as described above. Our estimates are very conservative and under-estimate the power for all except for~\cite{puglia2017real}.

\stepcounter{ti}
\begin{table*}[bhtp!]
    \centering
    \resizebox{\linewidth}{!}{    \scriptsize
    \centering
    \begin{tabular}{
        p{2.2cm}  
        >{\centering\arraybackslash}p{0.5cm}    
        >{\centering\arraybackslash}p{1.1cm}    
        >{\centering\arraybackslash}p{1.2cm}    
        >{\centering\arraybackslash}p{2cm}    
        >{\centering\arraybackslash}p{1.2cm}    
        >{\centering\arraybackslash}p{1cm}      
        >{\centering\arraybackslash}p{0.5cm}    
        >{\centering\arraybackslash}p{0.5cm}    
        >{\centering\arraybackslash}p{2.3cm}    
        >{\centering\arraybackslash}p{0.55cm}    
        >{\centering\arraybackslash}p{0.55cm}    
        >{\centering\arraybackslash}p{0.55cm}    
        >{\centering\arraybackslash}p{0.55cm}    
    }
        \toprule
        \textbf{Name} & \textbf{Type}       & \textbf{FPGA}              & \textbf{Resolution} & \textbf{Stereo Disparity Levels} & \textbf{Mpix/sec} &
        \multicolumn{3}{c}{\textbf{Estimated Core Power (W)}} & \textbf{Measured Full-System Power} & 
        \multicolumn{4}{c}{\textbf{Resource Utilization}} \\ 
        & & & & & & Reported & Min & Max & & LUTs & REGs & BRAMs & DSPs\\
        \midrule
        
        Jin, 2014 \cite{jin2014fast}    & Stereo            & Kintex-6                & 640$\times$480         & 60 & 15.6 &
        10.6 & 1.93 & 2.67 & none & 61000 & 61000 & 165 & 0\\

        Raj, 2014 \cite{joseph2014video}  & DfD        & Virtex-4            & 400$\times$400        & n/a & 48.0 &
        none & 0.46 & 0.58   & 2W + camera & 10028 & 7609 & 29 & 50   \\

        Mattoccia, 2015 \cite{mattoccia2015passive} & Stereo          & Spartan-7               & 640$\times$480   & 32 & 9.2 &
        none &  0.44& 0.68 & 2.5W & 23749 & 9030 & 67 & 0 \\

        Ttofis, 2015 \cite{ttofis2015low}    & Stereo            &  Kintex-7               &  1280$\times$720        & 64 & 55.3 & 
        2.8 & 0.92& 1.53 & none & 57492 & 71192 & 302 & 458 \\
        
        Puglia, 2017 \cite{puglia2017real}   & Stereo            &  Zynq Artix             &  1024$\times$768       & 64 & 23.6 & 
        0.17 & \underline{0.43} & \underline{0.68} & \underline{2W} & 29057 & 15208 & 80 & 0 \\

        Zhang, 2018 \cite{zhang2018nipm}    & Stereo            & Kintex-7                & 1920$\times$1080        & 128 & 124.4 &
        none & 0.89& 1.23 & none & 53190 & 40980 & 151 & 0 \\

        Lu, 2021 \cite{lu2021resource}    & Stereo            & Kintex-7                & 1024$\times$480        & 128 & 57.0 &
        none & 1.39& 2.30 & none & 50465 & 48046 & 125.5 & 8 \\

        Wang, 2022 \cite{wang2022block}    & Stereo            & Zynq US+                & 1920$\times$1080       & 256 & 344.2 &
        3.35 & 2.03& 2.20 & 2.8W + camera & 92300 & 59300 & 332.5 & 0 \\
        
        Ours, on Kintex-7                     & DfD       & Kintex-7            & 512$\times$480          & n/a & 59.9 &
        - & \textbf{0.42}& \textbf{0.55} & - &  24173 & 21947 & 58 & 58 \\
        
         Ours, actual, on ECP5                     & DfD       &  ECP5            & 480$\times$400          & n/a & 30.5 &
        - & \textbf{0.24} & \textbf{0.31}  & \textbf{0.6W} & 47339\textsuperscript{\textdagger} & 35107\textsuperscript{\textdagger} & 95\textsuperscript{\textdagger} & 67\textsuperscript{\textdagger} \\
        
        \bottomrule
    \end{tabular}}
    \caption{\textbf{Extended version of~\cref{tab:powertable}.} This table includes the reported power from other works. We re-did calculations from other works to account for differing pixel processing rates - our analysis favors methods that process pixels above 30 Mpix/s and penalize works below 30 Mpix/s. We under-estimate power for all works except~\cite{puglia2017real}. \\
    \textsuperscript{\textdagger} The internal design of the Lattice ECP5 means that its resource utilization values can't be directly compared to those of other FPGAs. Each LUT, BRAM, and DSP module on the ECP5 is smaller than on the 7-Series, so a design on the ECP5 will typically take more of these smaller resources.}
    \label{tab:detailed_powertable}
\end{table*}

\stepcounter{fi}
\begin{figure*}
    \centering
    \includegraphics[width=\linewidth]{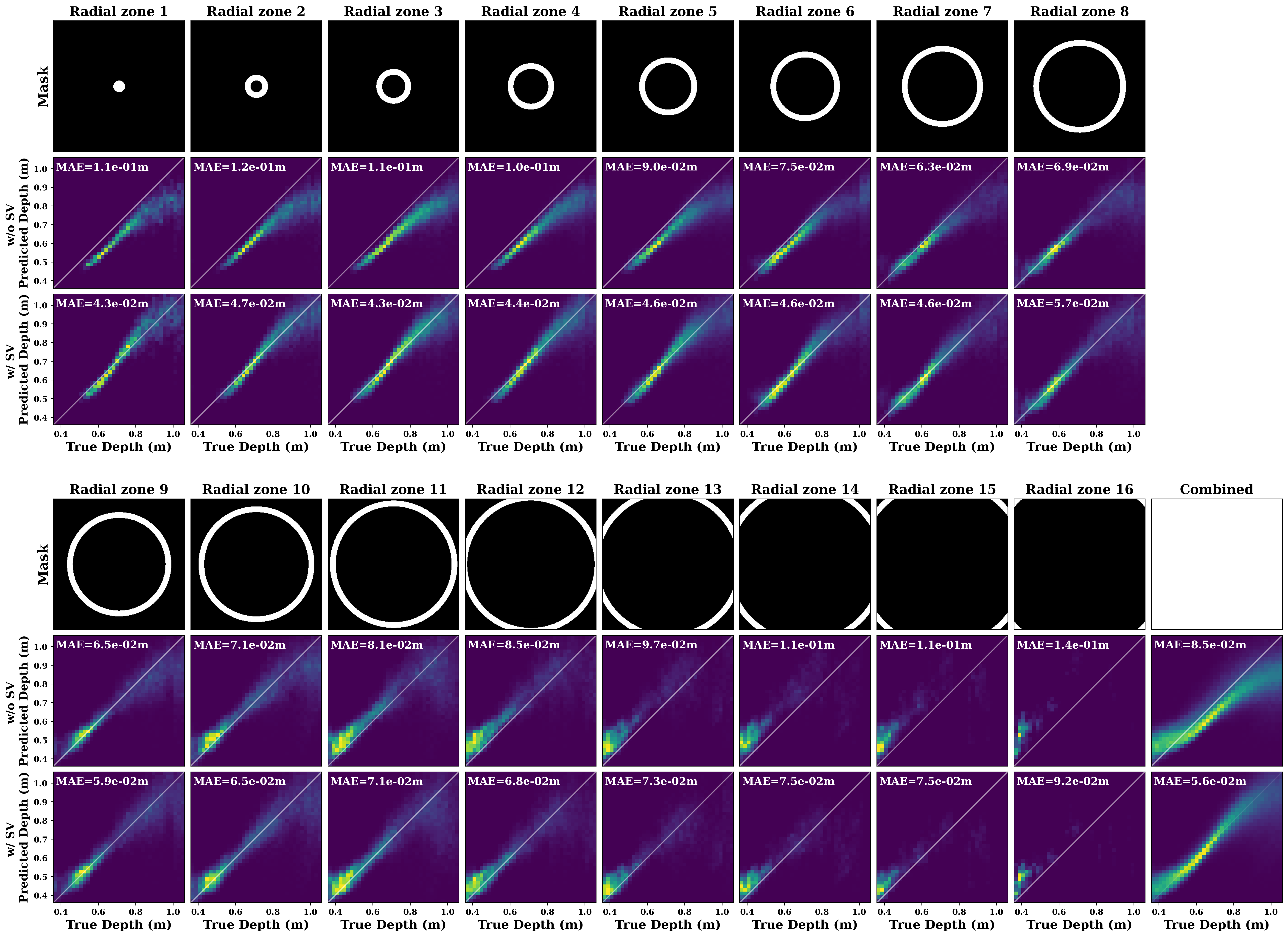}
    \caption{
        {\bf Heatmap comparison of our method with and without field curvature correction.}
        We decompose an image in 16 radial regions (row 1) and show the predicted depth vs. true depth heatmaps (filtered with a 90\% sparsity) for every radial regions and their combined region.
        We show that compared to using constant optical parameters (row 2), applying spatial variation (row 3) yields better depth predictions that are much more aligned with the diagonal, indicating better accuracy, in all radial zones.
        We also observe with the confidence filtering, radial zones in the center tend to perform better at longer depth ranges than the peripheral zones, which is well aligned with the observation of Petzval field curvature, where the focal planes are curved towards the lens.
    }
    \label{fig:heatmap_rings}
\end{figure*}

\stepcounter{fi}
\begin{figure*}
    \centering
    \includegraphics[width=\linewidth]{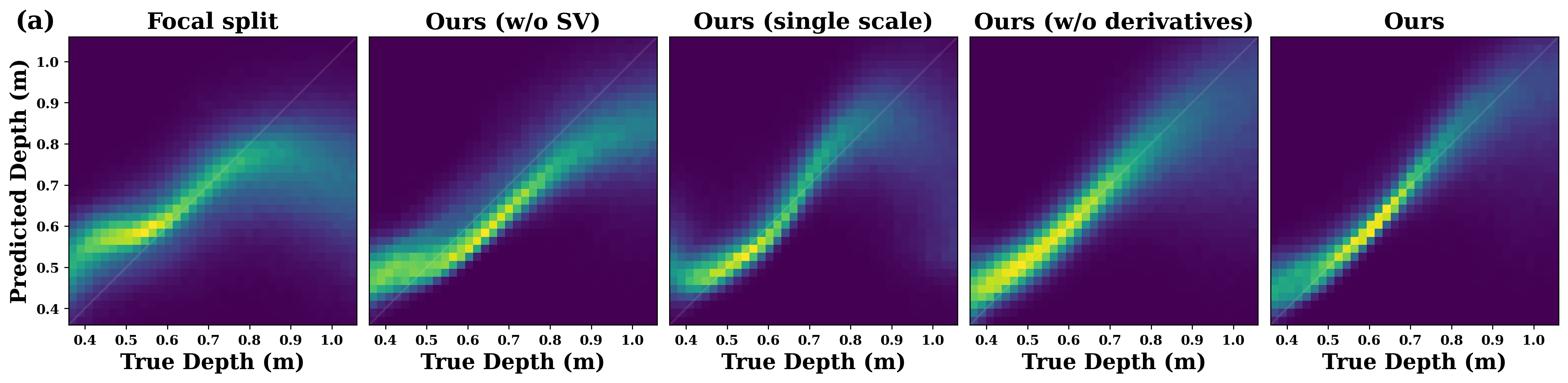}
    \includegraphics[width=\linewidth]{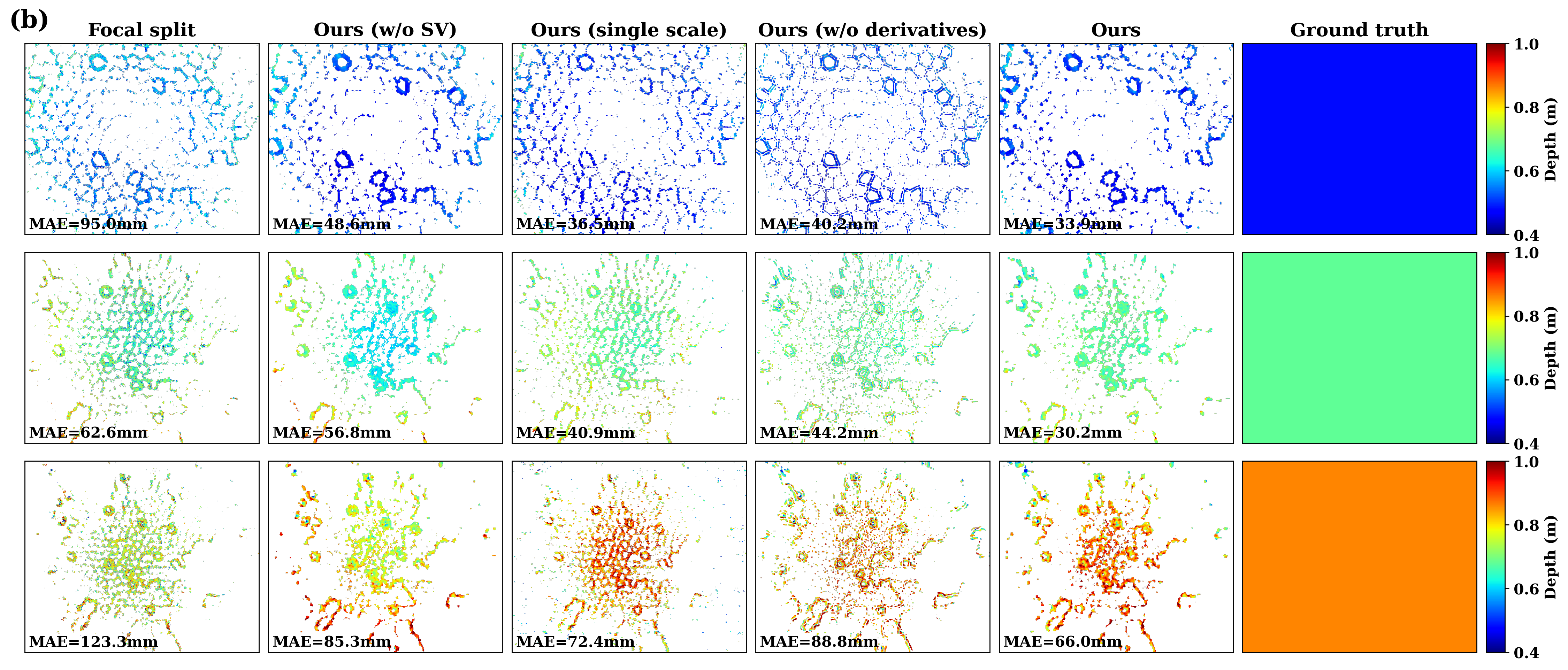}
    \caption{
        {\bf Ablation studies.}
        (a) Heatmaps of Focal Split~\cite{luo2025focal} and different variations of our method.
        (b) Predicted depth maps on the calibration dataset at three different depths from our front-parallel calibration dataset.
        We show that incorporating spatial variation, multiscale, and multi-derivative in our algorithm helps improve depth estimation in our lower power setting.
    }
    \label{fig:ablations}
\end{figure*}

\stepcounter{fi}
\begin{figure}
    \centering
    \includegraphics[width=\linewidth]{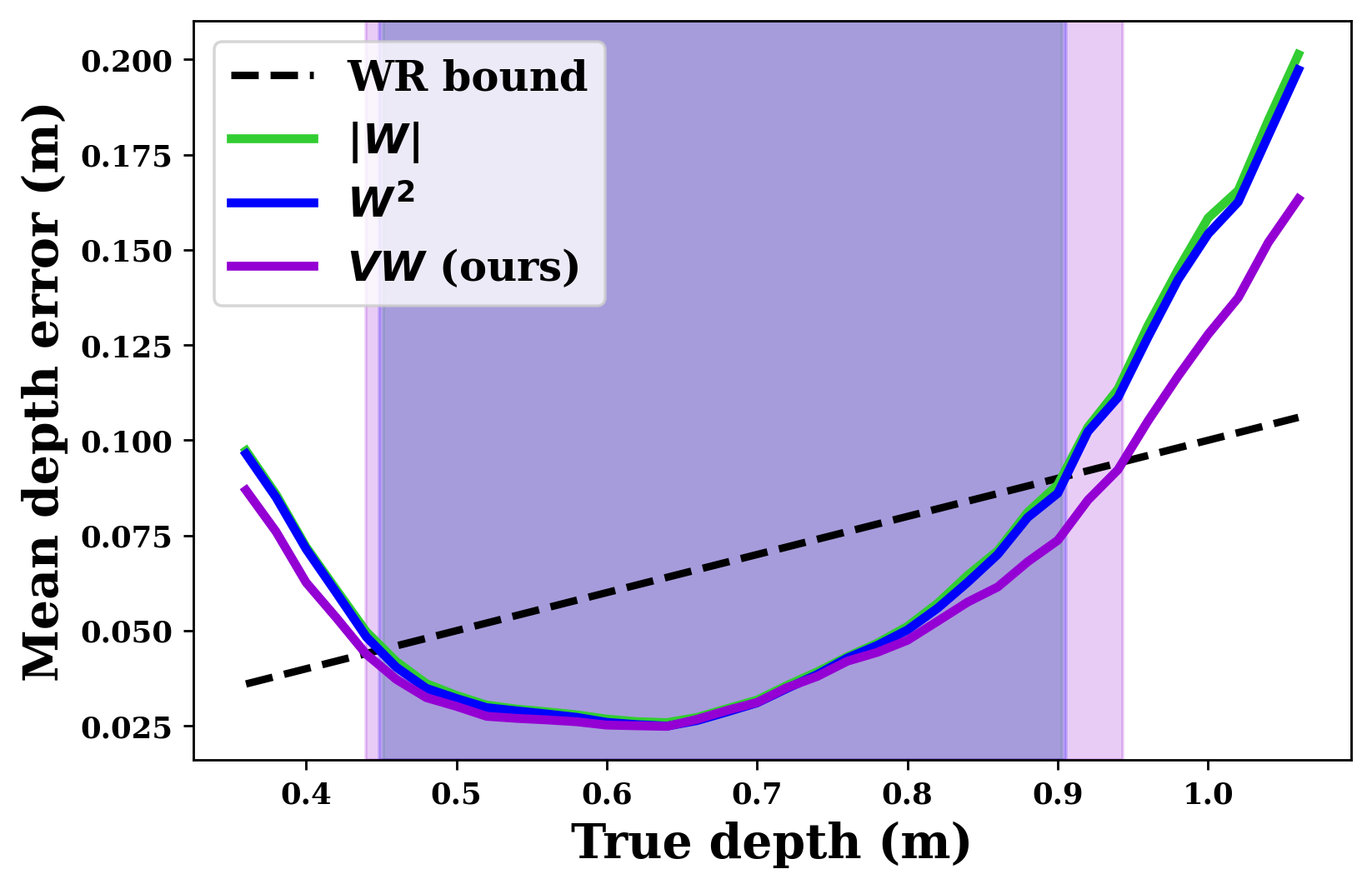}
    \caption{
        {\bf Comparison of different confidence.}
        We show the mean depth error vs. true depth curves at 90\% sparsity on the calibration data using different confidence metrics.
        We observed an increase in the working range with $VW$ (numerator in~\cref{eq:jointZ}) as confidence compared with $W^2$ or $\vert W \vert$, which were used in previous work~\cite{luo2025focal,luo2025depth}.
    }
    \label{fig:conf}
\end{figure}

\stepcounter{si}
\section{DIY Guide}
Except for a few custom PCBs and 3D printed optomechanical parts, SpiderCam is entirely made up of commercial off-the-shelf components. The custom components can be cheaply and readily manufactured using PCB designs and 3D CAD files provided by us. Consult~\cref{tab:partslist} for a list of all parts and how they can be acquired.

\stepcounter{ti}
\begin{table*}[t]
    \centering
    \smaller
    \label{tab:parts}
    \begin{tabular}{l|c|l|c}
        \toprule
        \textbf{Component Description} & \textbf{Commercial off-the-shelf} & \textbf{Part number / path to design files on GitHub} & \textbf{Qty} \\
        \midrule
        FPGA development board & yes & LFE5UM5G-85F-EVN & 1 \\
        Image sensor cables & yes & 900-0150180325-ND & 2 \\
        FT232H breakout board & yes & 2264 (Adafruit) & 1 \\
        16mm beamsplitter & yes & CCM5-BS016 & 1 \\
        M09 mounted lens & yes & AC050-008-A-ML & 2\\
        Lens adapter & yes & S05TM09 & 2\\
        Image sensor boards & no & \path{pcb/ecp5-fpc-adapter} & 2 \\
        Image sensor adapter & no & \path{pcb/hm0360-carrier} & 1 \\
        FT232H adapter board & no & \path{pcb/ft232h-breakout} & 1 \\
        3D printed optics fixture & no & \path{3d_printed_parts/dual_sensor_enclosure} & 2 \\
    \bottomrule
    \end{tabular}
    \caption{A list of all of the significant parts making up our system. 
    Commercial off-the-shelf components can be purchased at a store, while
    non-commercial off-the-shelf parts are those which we designed for this project and can be found in our GitHub repository (\href{https://github.com/NUBIVlab/SpiderCam}{\url{https://github.com/NUBIVlab/SpiderCam}}) using the path names. 
    To replicate our work, our design files must be used to order these from manufacturers.}
    \label{tab:partslist}
\end{table*}

\subsection{Electrical components}
\label{sec:diy_electrical_supp}

Our system uses a Lattice ECP5 FPGA for its processing. We deploy our FPGA Gateware on a Lattice ECP5 Evaluation Board, part number LFE5UM5G-85F-EVN. This board carries a Lattice LFE5UM5G-85F FPGA as well as a large amount of support circuitry and peripherals. Because this evaluation board is not tailored to our specific application, we slightly modified it to reduce power consumption. We removed an unused oscillator (refdes X2), a number of unused LEDs (refdes D22-26, D31), and removed inefficient linear regulators (refdes U5-6) supplying power to unused transceivers. We used a manually soldered wire to power the transceivers from the 1.2V switching regulator (refdes U8). While these modifications save some power, they don't alter the fundamental low-power contributions: more time and a larger budget would permit the construction of a project-specific PCB (as in~\cite{mattoccia2015passive}) instead of using a modified off-the-shelf board. An entirely custom PCB would admit even greater power savings. Furthermore, to make our system work, we had to remove R99 and add a 0-$\Omega$ resistor to footprint R104 to change VCCIO6 from 3.3 V to 2.5 V.

Although the low-power image sensor that we chose for this project can be readily purchased, at the time of writing, no PCBs carrying this image sensor without integrated optics are available. To this end, we constructed a carrier board for the HM0360 image sensor which we integrated into our optomechanical assembly. The HM0360 carrier board connects to the FPGA evaluation board through a 30-pin flat-flex cable and a custom-fabricated adapter board which connects two image sensors to the FPGA evaluation board through some of its header pins. Finally, data readout is achieved via another adapter board which connects an FT232H USB-to-Parallel transceiver to header pins on the FPGA board.

Construction of all of these PCBs costs only a few hundred dollars, and copies can be ordered fully assembled from many PCB fabrication providers. KiCAD design files, which can be used to order copies, can be found at \href{https://github.com/NUBIVlab/SpiderCam/tree/main/pcb}{\url{https://github.com/NUBIVlab/SpiderCam/tree/main/pcb}}.

\textbf{Assembling the system} Two image sensor boards affixed to the optical setup, as described in~\cref{sec:diy_optomechanical_supp}, need to be connected to the image sensor adapter board via flat-flex cables. The image sensor adapter board should be plugged into the FPGA development board via headers J39 and J40. An FT232H breakout board should be connected to the FPGA development board via headers J5 and J8 with the FT232H breakout adapter. A photograph of a fully assembled system can be seen in~\cref{fig:method}.

The FPGA board must be programmed with the Gateware that we developed for this project. \texttt{openocd} can be used to program the board with the appropriate Gateware images; prebuilt Gateware images can be found at \href{https://github.com/NUBIVlab/SpiderCam/tree/main/fpga/synth/dual_scale}{\url{https://github.com/NUBIVlab/SpiderCam/tree/main/fpga/synth/dual_scale}} along with programming instructions. Once the FPGA has been programmed, Python scripts at \href{https://github.com/NUBIVlab/SpiderCam/tree/main/fpga/tools/serialcam_ft232h}{\url{https://github.com/NUBIVlab/SpiderCam/tree/main/fpga/tools/serialcam_ft232h}} can be used to read data from the board using the FT232H USB-to-Parallel adapter. Explanation of the directory structure of different variants of the bit files and usage of the Python GUI is detailed at \href{https://github.com/NUBIVlab/SpiderCam/blob/main/fpga/README.md}{\url{https://github.com/NUBIVlab/SpiderCam/blob/main/fpga/README.md}}.

\subsection{Optomechanical assembly}
\label{sec:diy_optomechanical_supp}

\begin{figure}
    \centering
    \includegraphics[width=0.8\linewidth]{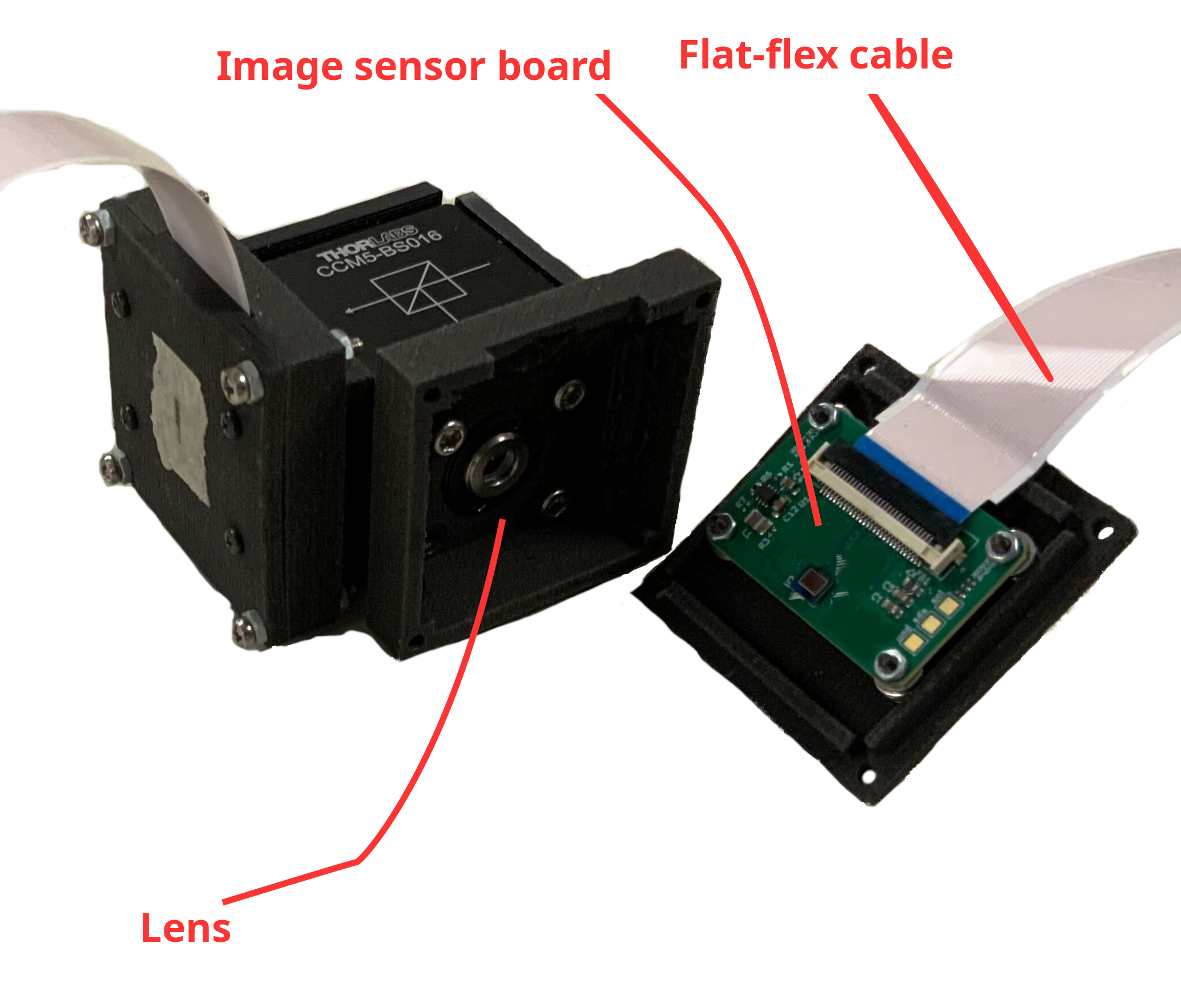}
    \caption{\textbf{Optical assembly.} This photograph shows the 3D printed mount that affixes the image sensor to the beamsplitter and aligns it to the lens. The image sensor, lens, and flat-flex cable connecting the optical assembly to the FPGA development board are visible.}
    \label{fig:optics_assembly}
\end{figure}

As described in~\cref{sec:camera}, our system relies on taking 2 pictures of the same scene with different levels of focus. To achieve this, we use a Thorlabs CCM5-BS016 beamsplitter with a 10mm AC050-008-A-ML lens attached to each output port. Each of the HM0360 image sensor boards is aligned with the optics using a 3D printed mount; the HM0360 image sensor boards are affixed to the mounts with M1.6 bolts, and the mounts are bolted onto the beamsplitter, aligning the two image sensors with their respective lenses. One side of a partially assembled image sensor mount can be seen in~\cref{fig:optics_assembly}.

\stepcounter{fi}

\textbf{Optomechanical assembly calibration} In order to achieve the defocus effect, the focal distance of the two sensors must vary. We used precision shims to offset the image sensors in the direction of the lenses optical axes and screwed the lenses in by different amounts to achieve a more precise offset.

A homography calibration was also needed. The pixel pitch on our image sensors is 3.6 micrometers, but the 3D printed optical mounts are manufactured with a tolerance of about 200 micrometers. Because of this, the images in the 2 image sensors are offset. After assembling the system, we manually calculate a homography; at runtime, we apply it to both images to align them before processing as described in~\cref{sec:algorithm}. To calculate the homography, we capture several photographs of a checkerboard pattern and manually align them using a custom tool found at \href{https://github.com/johnMamish/manual-homography-gui}{\url{https://github.com/johnMamish/manual-homography-gui}}. To capture the photographs for homography, the FPGA board must be reprogrammed with the Gateware image at \href{https://github.com/NUBIVlab/SpiderCam/tree/main/fpga/synth/camera_previewer}{\url{https://github.com/NUBIVlab/SpiderCam/tree/main/fpga/synth/camera_previewer}}.

\stepcounter{si}
\section{Depth Calibration}
\label{sec:suppcalibration}

We calibrate our prototype with experimentally collected front parallel texture plans at known depths. 
We collect 56 pairs of images of the textured plane at a depth range of 0.24m to 1.36m, with a step size of 0.02 m. 
We calculate the mean absolute error (MAE) between the depth predictions of all image pairs and the true depth as the loss function for calibration.
We only calculate the MAE loss on the 10\% most confident pixels in each radial zone (see~\cref{sec:spatialvar}).
We observed that by doing so, we prevent the optical constants from being biased by the noisy pixels and yield better depth predictions.
We use the AdamW optimizer with a learning rate of 0.05 and optimize for 100 iterations.
More details about our optimization parameters can be found in the configuration files within our GitHub repository.
We save our optimized optical constants $a,b,\omega$ in~\cref{eq:Z,eq:jointZ} in the supplementary numpy files.

During inference, we set constant confidence thresholds for each radial zone to ensure reliable depth predictions.
We use a log uniform confidence threshold from 1$\times$10$^{-7}$ (radial zone 1) to 2$\times$10$^{-5}$ (radial zone 16).
The peripheral radial zones have larger thresholds because they are more affected by field curvature and aberrations.
The starting point and ending point of the thresholds are tuned to maximize the working range while maintaining a density of 5\% within the working range.

In~\cref{fig:heatmap_rings}, we show heatmaps (with enforced 90\% overall sparsity) in the 16 radial zones with and without spatial variation.
We observe that when the optical constants are universal (row 2), the central and peripheral radial zones exhibit strong biases in the depth prediction, with the depth prediction being smaller in the central zones and larger in the periphery.
This agrees with the fact that the focal plane is curved towards the lens at the periphery of the field of view due to Petzval field curvature~\cite{hecht2017optics}.
When spatially-varying optical constants are used, depth predictions are concentrated on the diagonals and the bias is no longer visible, which indicates much higher accuracy.
We also compare several design choices in our method with Focal Split~\cite{luo2025focal} as an ablation study (all results are shown with an enforced 90\% sparsity in all depth maps).
In the heat map comparison in~\cref{fig:ablations}a, we show our depth predictions are much more concentrated on the diagonal compared to Focal Split, ours without spatial variation, ours with a single scale, and ours without derivatives.
In~\cref{fig:ablations}b, we compare predicted depth maps on three image pairs at three different depths in our calibration dataset.
We show that our method robustly outperforms others with better accuracy, especially in the challenging peripheral areas.
These are strong evidences that the incorporation of spatial variation, dual-scale, and spatial derivatives is improving depth predictions.
In~\cref{fig:conf}, we compare three confidence metrics: $|W|$ used in previous DfDD methods, and $W^2$ and $VW$, which are available without extra computation due to the depth computation. 
We show that using $VW$ (numerator in~\cref{eq:jointZ}) yields the widest working range with a fixed 90\% sparsity (45.1 cm for $|W|$, 45.7 cm for $W^2$, 50.3 cm for $VW$).

\stepcounter{si}
\section{Additional scenes and video}

\begin{figure*}[ht]
    \centering
    \includegraphics[width=1\linewidth]{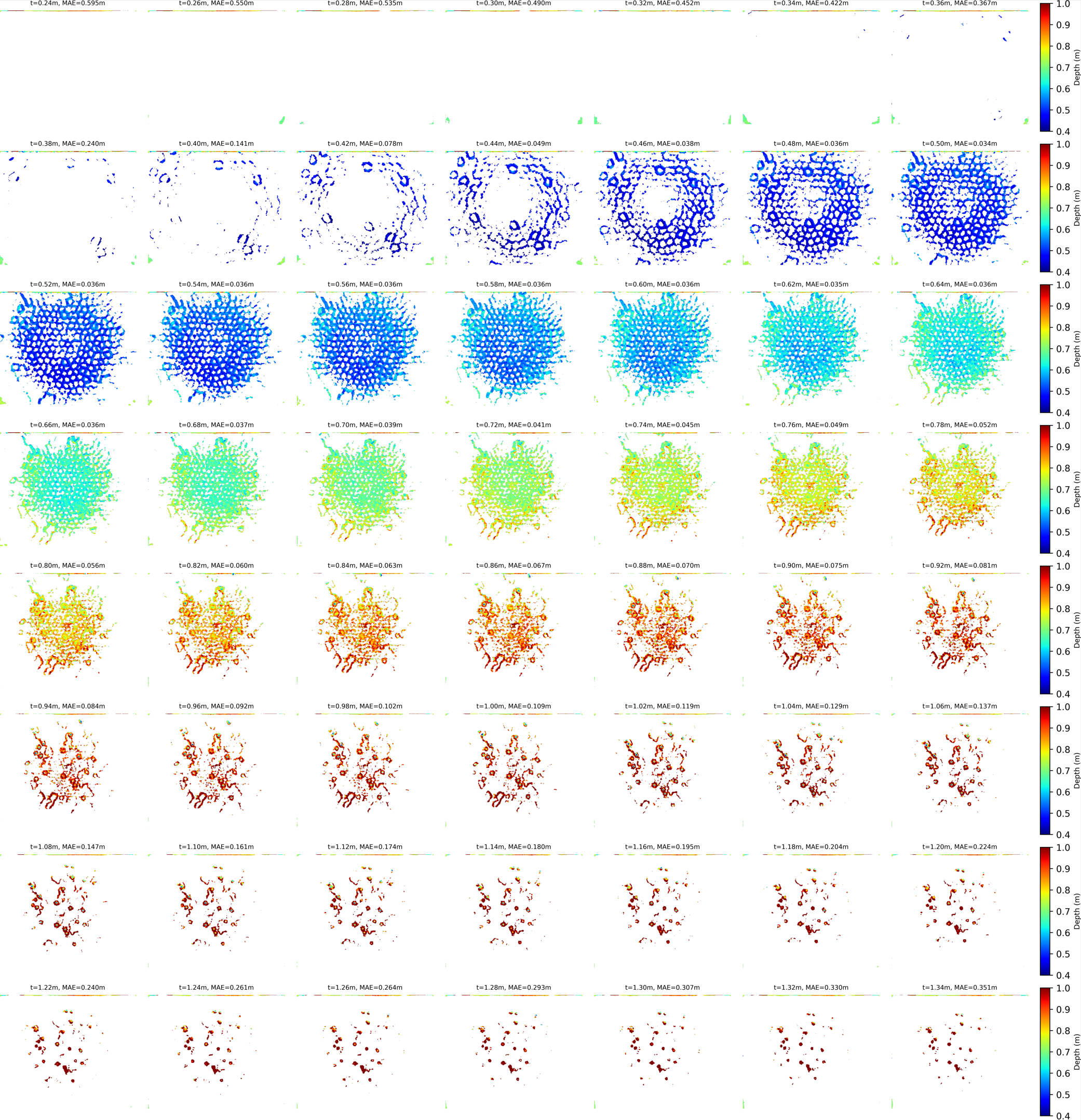}
    \caption{\textbf{Depth maps computed on-device for front-parallel textured plane.} This data was used to generate purple dots in \cref{fig:quant_analysis}. Note that density naturally falls off away from the center of the working range. Our confidence estimate captures edge textures with sufficient focus, and field curvature shifts the effective working range across the image. Edge effects dominate error in sparse images.}
    \label{fig:on_device_front_parallel}
\end{figure*}

\begin{figure*}[ht]
    \centering
    \includegraphics[width=1\linewidth]{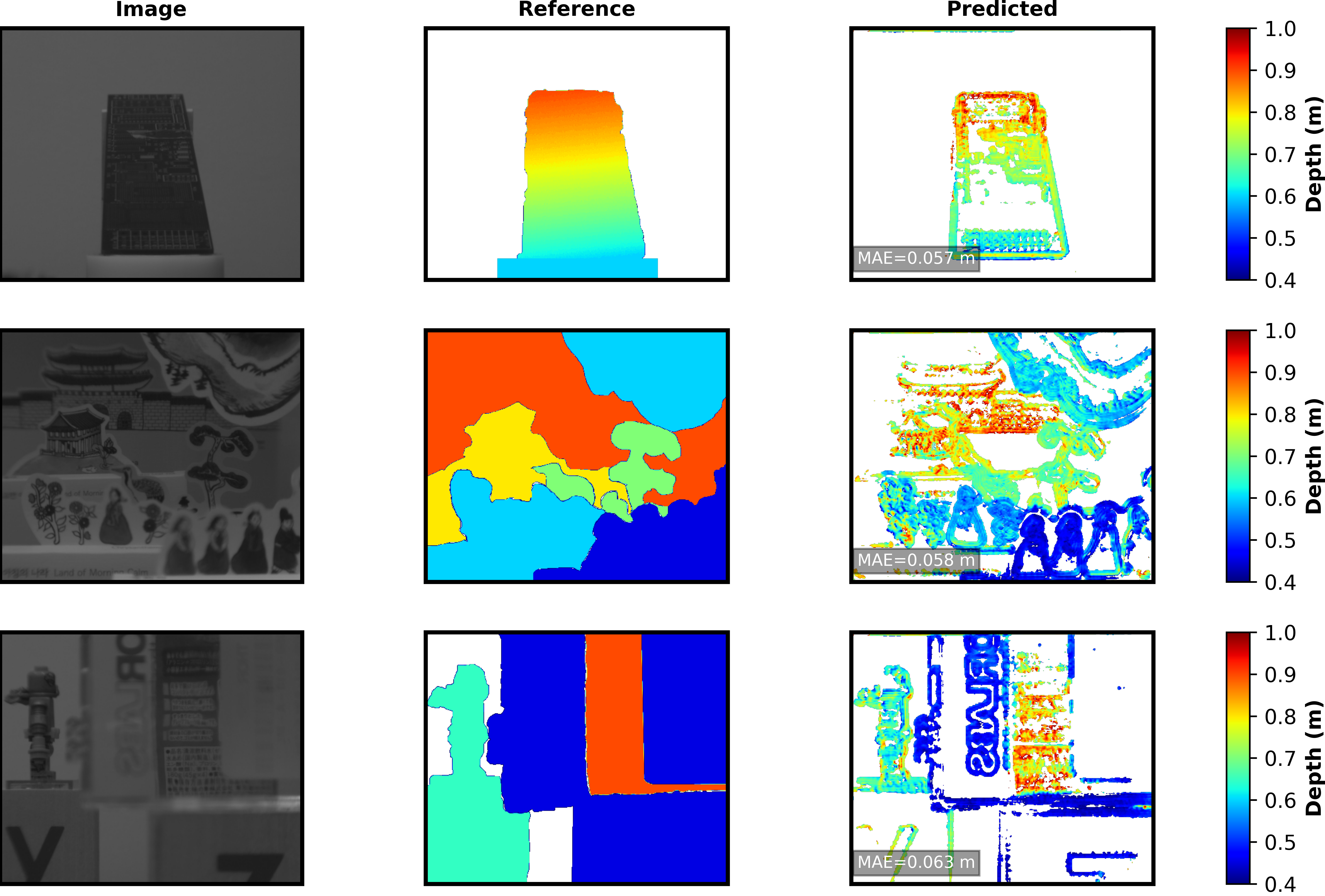}
    \caption{\textbf{Depth maps computed on-device with their references for real scenes.} Note MAE from top to bottom of 0.057 m, 0.058 m, and 0.063 m, consistent with data in \cref{fig:on_device_front_parallel}}
    \label{fig:on_device_real}
\end{figure*}

\noindent
See full display of the depth maps produced on device using the front-parallel textures and their respective MAE at \cref{fig:on_device_front_parallel}. This is the data that is concisely plotted as purple dots on \cref{fig:quant_analysis}. Furthermore, to show that the MAE from the font-parallel textures generalizes well, view \cref{fig:on_device_real} to see the MAE across additional real scenes. Generating the reference maps was done by placing object planes at known distances, then manually tracing the outlines and filling them in with appropriate values (or a gradient of values). Finally, we have the demo video titled 'demo.mp4' in the supplementary material, demonstrating real-time 32.5 FPS performance of our algorithm in action. 

\stepcounter{fi}

\stepcounter{fi}

\stepcounter{si}
\section{Full System Power Evaluation}
\begin{figure*}
    \centering
    \includegraphics[width=0.9\linewidth]{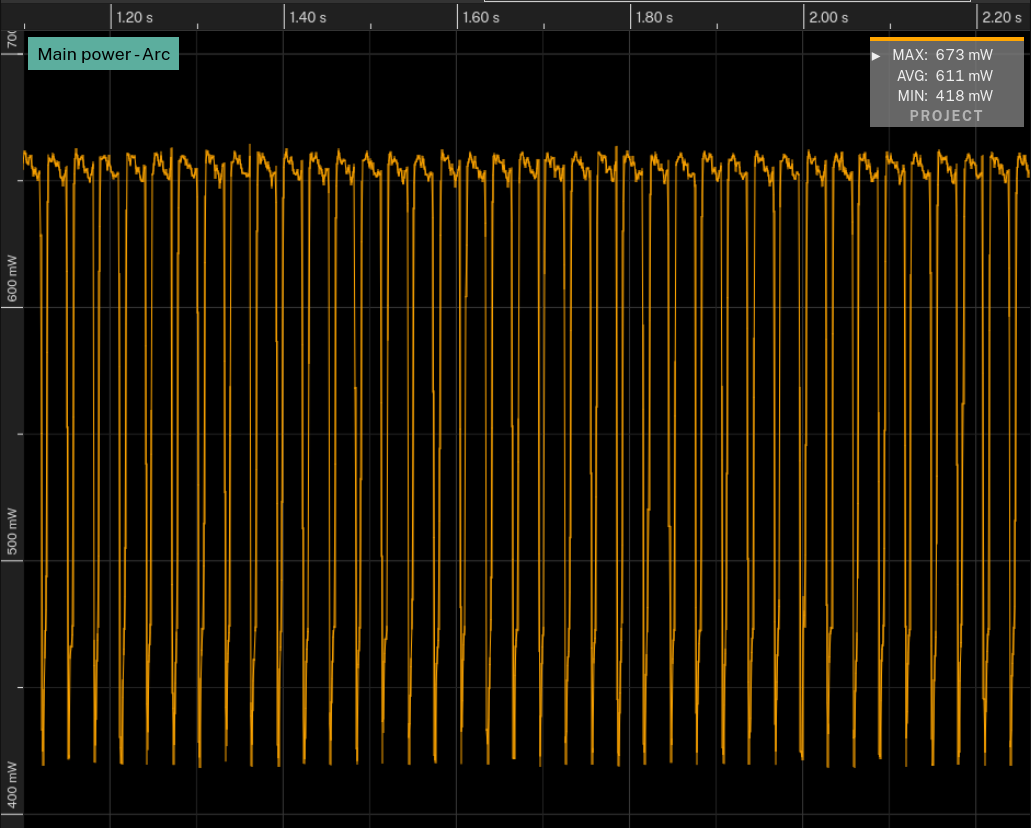}
    \caption{\textbf{Screenshot of representative power consumption trace.} We demonstrate how we used the Otii Arc Pro to measure power consumption for~\cref{tab:wallpower} by measuring power over time. This screenshot shows power for our system measured in a dual-scale configuration with dx and dy filters at 32.5 FPS. Notably, the frame rate is evident from the frequency of fluctuations in power consumption as seen in this graph - during active processing of a frame, power consumption is 650 mW; in-between frames, it drops to 420 mW. Our average power is 611 mW, which is slightly lower than our demonstrated 624 mW due to having captured this trace when the FPGA was cold compared to an extended period of usage during the recording of our demonstration that allowed the chip to warm up to its stable temperature. At the higher warm temperature, resistive losses in the device rise as well, resulting in slightly higher overall power draw.}
    \label{fig:power_trace_screenshot}
\end{figure*}

We measured the power consumption of our system in several configurations, shown in~\cref{tab:wallpower}. To perform the power measurement, we used a Qoitech Otii Arc Pro DC energy analyzer. The Otii Arc Pro supplied our system with 4.2 V of power and recorded current consumption at 4 ksps. We used Otii's software to calculate average power consumption over a 5-second window. A representative power trace can be seen in figure~\cref{fig:power_trace_screenshot}.

\stepcounter{fi}

During power measurements, the system calculated depth for real scenes similar to those in~\cref{fig:real_scenes}. To verify system functionality during measurement, we read data off as normal using a USB-to-Parallel converter and monitored data output. We designed the USB adapter board so that it cannot supply power to our system, but to rule out parasitic power supply via GPIO pins, we took power measurements with and without the USB adapter board attached and found that there was no noticeable difference.

\stepcounter{si}
\section{Power / Accuracy Trade-Off Details}

\stepcounter{ti}
\begin{table}
    \caption{\textbf{Full system power consumption} including image sensors, I/O, and FPGA, as reported in the literature and measured for several configurations of our method.}
    \label{tab:wallpower}
    \smaller
    \begin{tabular}{cccrr}
        \toprule
        \textbf{Work} & \textbf{Scales} & \textbf{Compute $I_x, I_y$}  & \textbf{FPS}   & \textbf{Overall power (mW)} \\ 
        \midrule
        \cite{luo2025focal}     & - & -   & 2.1  & 4900 \\
        \cite{puglia2017real}   & - & -   & 30   & 2000   \\
        \cite{mattoccia2015passive} & - & - & $ > 30 $  & $< 2500 $ \\
        ours                    & 2 & yes & 32.5 & 624 \\
        ours                    & 2 & no & 32.5 & 562 \\
        ours                    & 1 & yes & 32.5 & 489 \\
        ours                    & 1 & no & 32.5 & 468 \\
        ours                    & 2 & yes & 9 & 399 \\
        ours                    & 2 & no & 9 & 387 \\
        ours                    & 1 & yes & 9 & 356 \\
        ours                    & 1 & no & 9 & 347 \\
    \bottomrule
    \end{tabular}
\end{table}

In this section, we provide details on the power-accuracy tradeoff comparisons in~\cref{fig:tradeoffs} (see summary of working range, MAE, and core power in~\cref{tab:tradeoff}). We tested the power consumption and accuracy of several configurations of our system - by reducing the size of some computations, we can save power and reduce circuit size, but accuracy suffers, as can be seen in \cref{tab:tradeoff}.

For all methods that use spatial variation (circles and stars), we start with the same log-uniform confidence threshold described in~\cref{sec:suppcalibration}, and scale the thresholds together for the widest working range.
For the other methods using a universal threshold (squares), we tune the threshold value for the widest working range. This is necessary because the number of estimates changes the overall range of possible confidence estimates. 
We calculate the working range and depth MAE over 0.4 m to 1.0 m using the pixels whose confidence is above the confidence threshold.
In our code base, we provide an interface for users to change these options (spatial variation, number of scales, and spatial derivatives) in our method through a configuration file.

We characterized the power consumption of each configuration we tested as described in~\cref{sec:power_estimates_supp}. We synthesized the SystemVerilog design for each different configuration and used vendor tools for the ECP5 to estimate the power consumption.

\stepcounter{ti}
\begin{table}
    \caption{\textbf{Power-accuracy tradeoff for alternate algorithms.}
        We report the working range, MAE, and core power in~\cref{tab:tradeoff}.
    }
    \label{tab:tradeoff}
    \small
    \begin{tabular}{cccccc}
        \toprule
        \textbf{Scales} & \textbf{$I_x, I_y$}  & \textbf{SV} & \textbf{WR (m)} & \textbf{MAE (m)} & \textbf{Core power (mW)} \\ 
        \midrule
        2 & yes & yes & 0.551 & 0.044 & 312 \\
        2 & yes & no & 0.311 & 0.058 & 280 \\
        2 & no & yes & 0.282 & 0.075 & 276 \\
        2 & no & no & 0.259 & 0.071 & 257 \\
        1 & yes & yes & 0.423 & 0.069 & 214 \\
        1 & yes & no & 0.210 & 0.076 & 195 \\
        1 & no & yes & 0.175 & 0.089 & 202 \\
        1 & no & no & 0.074 & 0.098 & 184 \\
    \bottomrule
    \end{tabular}
\end{table}
\end{document}